\documentclass[10pt,twocolumn,letterpaper]{article}

\usepackage[pagenumbers]{iccv} 
\usepackage[hang,flushmargin]{footmisc}

\definecolor{iccvblue}{rgb}{0.21,0.49,0.74}
\usepackage[pagebackref,breaklinks,colorlinks,allcolors=iccvblue]{hyperref}
\usepackage{kotex}
\newcommand{\sysname}{PromptDresser}
\def\confName{ICCV}
\def\confYear{2025}

\title{PromptDresser: Improving the Quality and Controllability of Virtual Try-On \\ via Generative Textual Prompt and Prompt-aware Mask}

\author{
Jeongho Kim$^\text{*}$ \quad\quad Hoiyeong Jin$^\text{*}$ \quad\quad Sunghyun Park \quad\quad Jaegul Choo\\
KAIST, Daejeon, South Korea\\
{\tt\small \{rlawjdghek, hy.jin, psh01087, jchoo\}@kaist.ac.kr}
}

\definecolor{darkgreen}{rgb}{0.0, 0.75, 0.0} 
\definecolor{lightblue}{rgb}{0.0, 0.5, 1.0} 

\begin{document}

\renewcommand{\thefootnote}{\fnsymbol{footnote}}
\footnotetext[1]{Equal contribution.}

\begin{figure}
\twocolumn[{
\renewcommand\twocolumn[1][]{#1}
\maketitle
\begin{center}
    \centering 
    \includegraphics[width=1.0\linewidth]{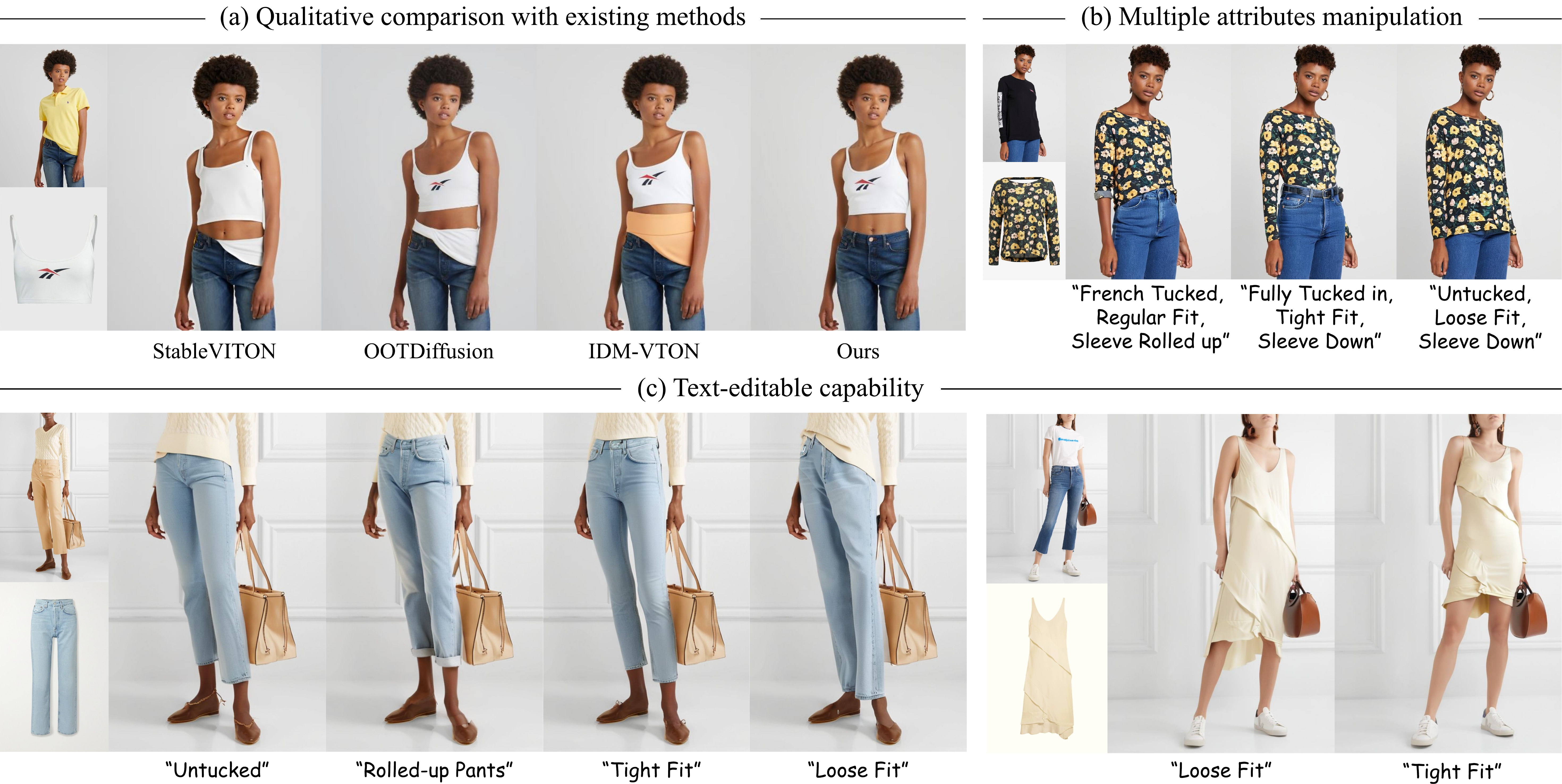}
    \vspace{-0.35cm}
    \captionof{figure}{Generated results of~\sysname: (a) effectively resolves the issue of residual clothing shape in existing methods (b) highlights multiple attributes manipulation simultaneously, and (c) generates versatile outputs across multiple clothing categories, including bottoms and dresses.}
    \vspace{-0.35cm}
    \label{fig:teaser}
\end{center}
}]
\end{figure}

\begin{abstract}
    Recent virtual try-on approaches have advanced by fine-tuning pre-trained text-to-image diffusion models to leverage their powerful generative ability; however, the use of text prompts in virtual try-on remains underexplored.
    This paper tackles a text-editable virtual try-on task that modifies the clothing based on the provided clothing image while editing the wearing style (e.g., tucking style, fit) according to the text descriptions.
    In the text-editable virtual try-on, three key aspects exist: (i) designing rich text descriptions for paired person-clothing data to train the model, (ii) addressing the conflicts where textual information of the existing person's clothing interferes the generation of the new clothing, and (iii) adaptively adjust the inpainting mask aligned with the text descriptions, ensuring proper editing areas while preserving the original person's appearance irrelevant to the new clothing.
    To address these aspects, we propose~\sysname, a text-editable virtual try-on model that leverages large multimodal model (LMM) assistance to enable high-quality and versatile manipulation based on generative text prompts.
    Our approach utilizes LMMs via in-context learning to generate detailed text descriptions for person and clothing images independently, including pose details and editing attributes using minimal human cost.
    Moreover, to ensure the editing areas, we adjust the inpainting mask depending on the text prompts adaptively.
    Our approach enhances text editability while effectively conveying clothing details that are difficult to capture through images alone, leading to improved image quality. 
    Experiments show that~\sysname~significantly outperforms baselines, demonstrating superior text-driven control and versatile clothing manipulation. Our code is available at \url{https://github.com/rlawjdghek/PromptDresser}.
\end{abstract}

\section{Introduction}
Inpainting-based Virtual Try-on~\cite{han2018viton} effectively preserves non-target clothing features (\textit{e.g.}, face or background), enabling personalized fashion previews by swapping items such as tops and bottoms.
Moreover, the capability to control wearing styles, such as tucking style or fit, can enhance overall user experience by allowing users to visualize different outfit options.

Thanks to the advanced generative capabilities of diffusion models~\cite{ho2020denoising,dhariwal2021diffusion,ramesh2021zero}, virtual try-on have shown significant improvements compared to earlier generative adversarial network-based approaches~\cite{goodfellow2020generative,wang2018toward,han2019clothflow,yang2020towards,choi2021viton,ge2021parser,lee2022high,bai2022single,gou2023taming}.
These recent models leverage the powerful priors embedded in large-scale diffusion models~\cite{rombach2022high,chen2023pixart,podell2023sdxl,saharia2022photorealistic}, resulting in improvements in clothing detail representation and generalization performance. 
Furthermore, recent studies have enhanced generative quality by incorporating text inputs with advanced text-to-image (T2I) models~\cite{choi2024improving,li2024anyfit} and have added an editing capability via click or drag~\cite{chen2024wear}. However, although these approaches are built on text-based generative models~\cite{rombach2022high,podell2023sdxl}, they typically use simple textual prompts of clothing attributes, such as `short sleeve t-shirt.' The potential for using rich textual information to achieve higher performance and extensive text-driven editability is still largely unexplored.

In this paper, we tackle a text-editable virtual try-on task that changes the clothing item based on the provided clothing image and edits the wearing style according to the text descriptions. 
Text-editable virtual try-on has unique and challenging aspects. 
(i) Constructing rich, yet compact and well-aligned text descriptions for dressing the person in new clothing is essential, (ii) while avoiding textual conflicts between the original and new clothing during sampling.
Since wearing styles (\textit{e.g.}, untucked, fully tucked in) should be adjustable through text, (iii) the mask must be adaptively change to align with the text prompt, preserving regions unrelated to clothing (\textit{e.g}., face or background) and removing the information of the original clothing shape.

To address these challenges, we introduce~\sysname, a novel virtual try-on model that leverages rich textual information to achieve high-quality and versatile outfit manipulation. 
To generate the text prompts for paired person-clothing data, we instruct the large multimodal models (LMMs) to describe the person and clothing images, respectively.
In virtual try-on, the target clothing differs from the originally worn clothing during inference. 
For a high-fidelity try-on, it is crucial to retain only agnostic information while removing details of the original clothing~\cite{choi2021viton}. 
Additionally, unmasked regions in the inpainting process already contain significant details, such as the face and non-target clothing.
Simply instructing an LMM to describe the image often produces mixed captions of person and clothing, or generate excessively redundant descriptions. 
Therefore, we leverage in-context learning that conditions expressive layouts, effectively specifying attributes to focus on the inpainting regions with minimal human cost.
We pre-define attributes for inpainting-based virtual try-on and provide human-labeled attribute examples for LMM in-context learning. 
This approach is significantly effective than holistic descriptions of the entire person and clothing, providing a scalable and efficient solution for virtual try-on. 
Moreover, we find that detailed text descriptions generated by LMMs not only improve text-based editability but also preserve intricate clothing details that images alone may not fully capture, resulting in high-fidelity virtual try-on images.

In virtual try-on \cite{choi2021viton, lee2022high, morelli2023ladi, kim2024stableviton, choi2024improving, chen2024wear}, inpainting masks play a crucial role in removing existing clothing details.
Notably, conventional masks often constrain the model to the original clothing shape, leading to unnatural results with different clothing types (see Figure~\ref{fig:teaser} (a)).
To address this, we introduce a novel masking technique, Prompt-aware Mask Generation (PMG), which adaptively adjusts according to the text prompt. 
We propose two types of mask: coarse and fine mask. 
The course mask is used to predict the form of the new clothing, agnostic to the existing clothing. 
It serves as a guide to indicate the areas needed for the new clothing to naturally generate while aligned with the text descriptions (e.g., outfit or tucking style). 
By combining the predicted mask and the fine mask, we adaptively mask out the existing clothing conditioned on the text prompt while preserve the non-target clothing region. 
To enable the model to learn from coarse to fine masks, we additionally applied random dilation mask augmentation during the training stage. 
This enables the model to learn how to adaptively generate across a wide range of mask regions, from broad to narrow, thereby improving its flexibility in handling various mask shapes. 

In summary, our contributions are as follows:
\begin{itemize}
    \item We propose a text-editable virtual try-on model that, with the assistance of a LMM with in-context learning, achieves rich, well-aligned text descriptions for both person and clothing, ensuring no textual conflicts with any clothing provided.
    \item To mitigate the issue of following the original clothing’s attributes, such as shape, we propose random dilation mask augmentation. Our prompt-aware mask generation enhances diversity in virtual try-on results while effectively preserving the person’s original appearance.
    \item Our approach achieves state-of-the-art performance across multiple datasets with and enables versatile manipulation capabilities, highlighting the effectiveness of generative textual prompt for virtual try-on.
\end{itemize}

\section{Related Work}
\subsection{Image-based Virtual Try-On}
Recent virtual try-on approaches build upon large-scale text-to-image diffusion models~\cite{saharia2022photorealistic,rombach2022high,podell2023sdxl,chen2023pixart}, inherently benefiting from their robust inpainting capabilities~\cite{gou2023taming} or utilizing textual inversion techniques~\cite{morelli2023ladi} or conditioning via attention mechanism~\cite{kim2024stableviton,choi2024improving}.
Notably, recent research~\cite{kim2024stableviton,choi2024improving} propose attention-based, end-to-end virtual try-on models that preserve fine details of clothing while achieving the generalization performance.

Another line of progress in image-based virtual try-on is controllability. LC-VTON~\cite{yao2023lc}, for instance, introduced new segmentation labels to incorporate clothing length, enabling the generation of high-fidelity images. 
Moreover, while spatial condition-based approaches~\cite{li2024controlling,chen2024wear} offer a fine degree of controllability, they are limited in their ability to perform comprehensive editing such as adjusting fit and overall appearance. 

On the other hand, recent research leveraging text-based controllability has tried to address such challenges, but fail to consider the existing areas (\textit{e.g.}, background) that need to be preserved~\cite{zhu2024m} or are limited to captioning solely for clothing~\cite{li2024anyfit,choi2024improving}.
While some methods condition on both text and clothing~\cite{shen2024imagdressing}, effective prompting and masking techniques for inpainting-based virtual try-on remain unexplored.
In this paper, we propose an effective prompting strategy for disentangling the person and clothing in inpainting-based virtual try-on, along with a novel adaptive masking method that enables more flexible manipulation while preserving the original person's appearance.

\subsection{LMMs for Multimodal Data Augmentation}
Recent advancements in large multimodal models (LMMs)~\cite{alayrac2022flamingo,touvron2023llama,liu2024visual,li2023blip,zhu2023minigpt} have demonstrated their powerful visual understanding~\cite{wang2024visionllm,luo2023valley}, achieving impressive performance across various vision tasks. Harnessing the capabilities of LMMs, they are utilized to tune image editing models~\cite{brooks2023instructpix2pix} and to enhance captioning performance though image-caption fusion~\cite{bianco2023improving}. However, their application to virtual try-on remains still underexplored. In this paper, we leverage off-the-shelf LMMs to augment image captions from virtual try-on datasets, enabling the generation of higher-fidelity images. Furthermore, we allow users to wear diverse styles through text-based manipulation according to their preferences.

\begin{figure*}[t!]
    \centering
    \includegraphics[width=\linewidth]{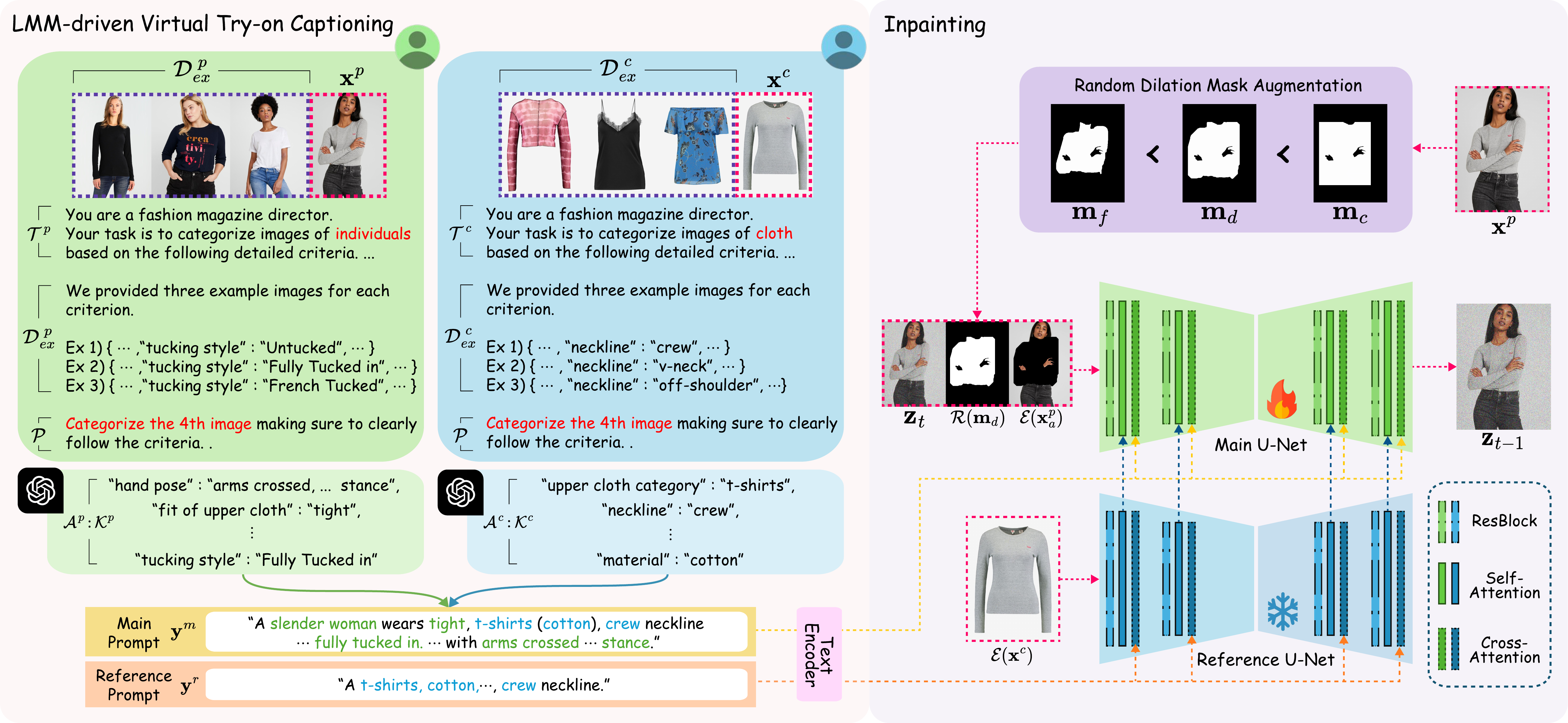}
    \caption{Overview of~\sysname. By using LMM with in-context learning, we generate two types of captions specific to the person and clothing images. The reference prompt consisting of clothing captions is provided to the reference U-Net, while a main prompt, including both clothing and person, is input into the main U-Net. To preserve the details of clothing, we use a frozen U-Net as a feature extractor. To enable learning across a diverse range of masked images, we randomly dilate the inpainting mask.}
    \label{fig:main}
\end{figure*}

\section{Method}
\subsection{Preliminary: Latent Diffusion Model}
Our approach builds on pre-trained text-to-image latent diffusion models (LDMs)~\cite{rombach2022high,podell2023sdxl}, which consist of three main components: a variational auto-encoder (VAE) with an encoder $\mathcal{E}(\cdot)$ and decoder $\mathcal{G}(\cdot)$, a text encoder $\tau(\cdot)$ and main U-Net $\epsilon_\theta(\cdot)$. The pre-trained VAE encodes an image $\mathbf{x}$ into a low-dimensional latent space as $\mathbf{z}_0=\mathcal{E}(\textbf{x})$ and reconstructs it back into RGB space as $\hat{\mathbf{x}}=\mathcal{G}(\mathbf{z}_0)$. The main U-Net is trained to predict the $\mathbf{z}_0$ from the perturbed latent variable $\mathbf{z}_t$, defined as $\mathbf{z}_t=\mathcal{N}\left(\mathbf{z}_t; \sqrt{\Bar{\alpha}_t}\mathbf{z}_0, (1-\Bar{\alpha}_t)\mathbf{I}\right)$. Here, $\Bar{\alpha}_t=\Pi_{s=1}^t(1-\beta_s)$, where $(\beta_t)_{t=0}^T$ is a decreasing sequence~\cite{ho2020denoising}.
The loss function of LDMs is given by:
\begin{equation}
    \mathcal{L}_{LDM} = \mathbb{E}_{\mathbf{z}_0\sim\mathcal{E}(\mathbf{x}),\epsilon\sim\mathcal{N}(0, I),\mathbf{y},t}\left[\| \epsilon - \epsilon_\theta(z_t,t,\tau(\mathbf{y})) \|_2^2\right].
\end{equation}
This objective function directs the model to reduce the difference between the added noise, $\epsilon$, and the noise predicted by the U-Net. The predicted noise is then used in the reverse diffusion process to approximate the original latent representation.

\subsection{Overall Framework}
Our method,~\sysname, aims to enable text-editable virtual try-on of reference clothing $\mathbf{x}^c$ onto a target person $\mathbf{x}^p$ with additional generative textual prompt provided by LMMs. An overview of the proposed method is illustrated in Fig.~\ref{fig:main}. Following the existing virtual try-on works~\cite{morelli2023ladi,gou2023taming,kim2024stableviton}, we adopt an inpainting framework, which reconstructs the target person image from a masked version, conditioned on the reference clothing image. Specifically, the main U-Net generates the target person image based on the input including a noise image ($\mathbf{z}_t$), a resized dilated clothing-agnostic mask ($\mathcal{R}(\mathbf{m}_d)$), and a latent agnostic map ($\mathcal{E}(\mathbf{x}^p_a)$). Here, the dilated clothing-agnostic mask $\mathbf{m}_d$ is produced by a random dilation mask augmentation.

To preserve fine clothing details, we leverage a frozen U-Net as a feature extractor~\cite{tang2023emergent,podell2023sdxl}, referred to as the reference U-Net. We then integrate the clothing features of reference U-Net by concatenating the key and value from self-attention layers of the reference U-Net with those of corresponding layers in the main U-Net~\cite{xu2024magicanimate}. 

To obtain rich text descriptions, we introduce an LMM-driven captioning mechanism.
Merely providing a detailed description on person images can result in redundancy, as it may encompass areas that remain unmasked. 
Furthermore, this approach fails to address the unique nature of unpaired person-clothing data in virtual try-on, where the information of the person’s original clothing and the new clothing can become entangled.
Therefore, we designed our approach to separate information by pre-defining distinct attributes for the person and clothing.

The existing clothing-agnostic person representation~\cite{choi2021viton,lee2022high} fails to precisely remove structural features like clothing length, causing the generated images to conform closely to the shape of the original clothing, which limits the flexibility for manipulation. On the other hand, overly expanded mask regions make it difficult to accurately reconstruct the original person's information. Therefore, we introduce random dilation mask augmentation, enabling the model to learn a range of mask images from coarse to fine. This approach allows for a prompt-aware mask generation (PMG) during the inference, providing preservation of the original person's appearance irrelevant to the reference clothing. 

\subsection{LMM-driven Virtual Try-on Captioning}
In this work, we propose to improve the quality and controllability of virtual try-on via generative textual prompt.
The primary challenges are: 1) \textit{ensuring that the text description focuses on the masked region} and 2) \textit{excluding textual information about the existing clothing in the inference process}.
A naive approach would be to instruct the LMM to describe the person image, which results in detailed information about visible features in the unmasked regions, such as expression and hairstyle, as in \textit{``The image showcases a bold fashion statement \ldots large hoop earrings and curly, voluminous hair enhance the overall stylish and confident look.''}. Moreover, describing the entire image in this way can introduce textual conflicts when virtually fitting new clothing, as information about the person’s existing outfit may also be included.

To address this, we propose designing pre-defined attributes and tasking the LMM with generating captions based on these attributes.
Using the LMM, we listed the attributes of both person and clothing images and carefully selected $n^{\{p,c\}}$ representative attributes $\mathcal{A}^{\{p,c\}} = \{a^{\{p,c\}}_1, \dots, a^{\{p,c\}}_{n^{\{p,c\}}}\}$. Specifically, for person-related attributes, we prioritized those within the masked region, including hand pose, body shape, and tucking style, along with attributes that support editability. Due to the limited token length of the backbone’s text encoder (\textit{i.e.}, CLIP), we focused on selecting global clothing features, such as category and material, rather than local details like logos.
The main advantages of using the LMM to generate captions based on pre-defined attributes for both person and clothing are: 1) it enables the generation of \textbf{informative text descriptions specifically for the inpainting areas} and 2) the separation of information for person and clothing allows us to \textbf{create adequate prompts aligned to unpaired person-clothing scenarios}, enabling descriptions of individuals wearing new clothing.
Furthermore, we instruct the LMM to provide a detailed description of the pose. By using pose descriptions instead of DensePose~\cite{guler2018densepose}, we retain the text-to-image backbone architecture and effectively reduce the errors associated with pose networks used in previous works, particularly in complex samples.

\begin{figure}[t!]
    \centering
    \includegraphics[width=\linewidth]{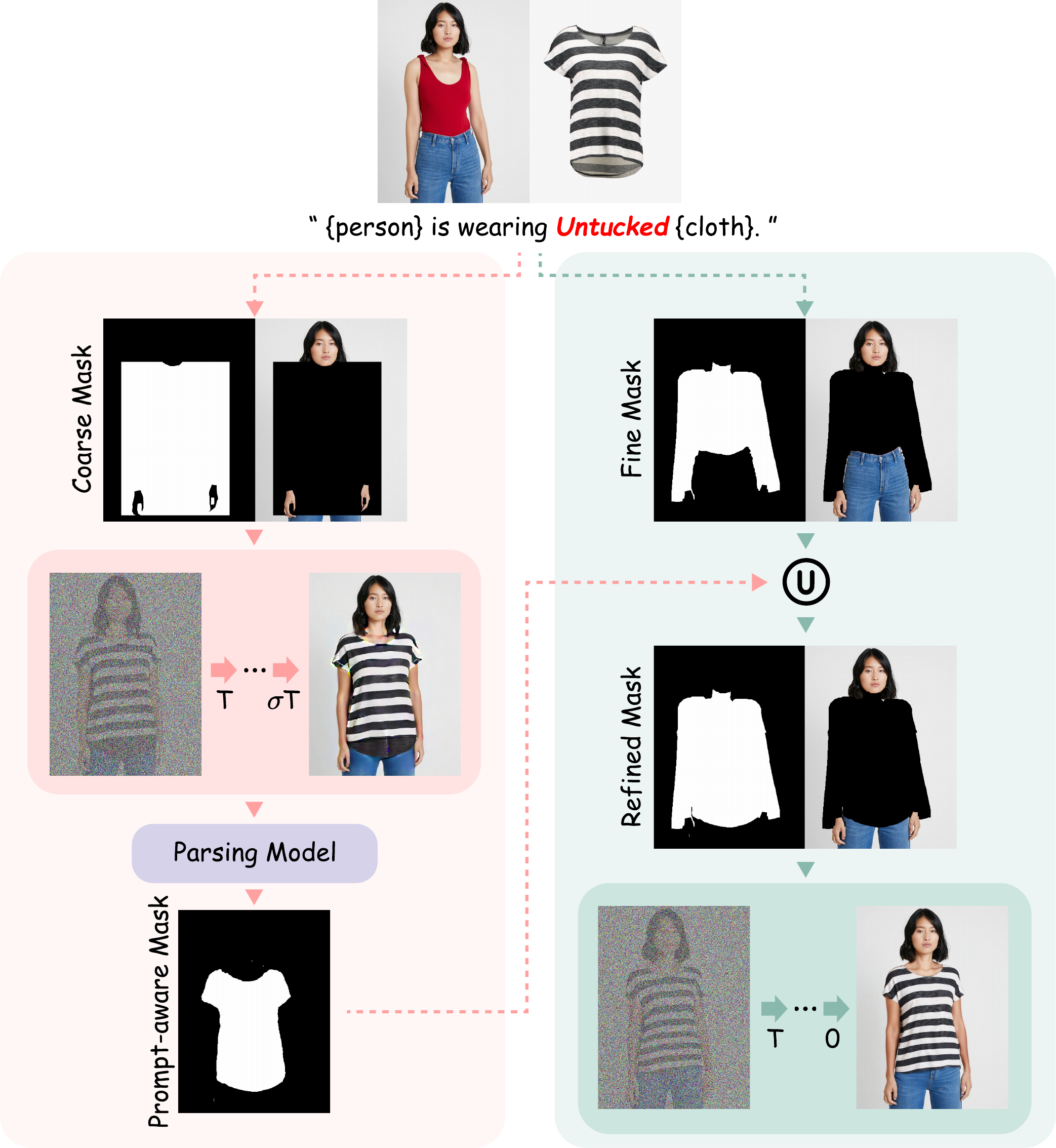}
    \caption{Prompt-aware mask generation for text-based manipulation. During the inference stage,~\sysname~takes a coarse mask as input and generates a prompt-aware mask. A refined mask for inpainting is then obtained by performing a union operation with a fine mask, ensuring minimal alteration to regions irrelevant to the clothing in the person image.}
    \label{fig:PMG}
\end{figure}

As depicted on the left of Fig.~\ref{fig:main}, we then utilize the in-context learning capability~\cite{brown2020language} of LMM models to generate rich, free-form captions for pre-defined attributes. Based on the observation that multi-modal models are proficient at predicting head categories such as gender~\cite{li2024cosmicman}, we carefully select $N$ few-shot exemplar images, each exhibiting different subtle details such as tucking or rolling style. Human annotators then label each caption to capture these specific details accurately. 

Therefore, for an arbitrary person or clothing image $\mathbf{x} \in \{\mathbf{x}^p, \mathbf{x}^c\}$ and a LMM $\mathcal{M}$, the predicted captions $\mathcal{K}^{\{p,c\}}$ is obtained through in-context learning as: 
\begin{equation}
    \mathcal{K}^{\{p,c\}}=\{k^{\{p,c\}}_1, \dots, k^{\{p,c\}}_{n^{\{p,c\}}}\}=\mathcal{M}(\mathbf{x} | \mathcal{P}, \mathcal{D}^{\{p,c\}}_{ex}, \mathcal{T}^{\{p,c\}}),
\end{equation}
where $\mathcal{P}$ is the input prompt, $\mathcal{D}^{\{p,c\}}_{ex}$ is the in-context learning dataset consisting of few-shot examples for captioning, and $\mathcal{T}^{\{p,c\}}$ is the task description. 

Using the predicted captions, we construct textual prompts for both the reference U-Net and the main U-Net. For the reference U-Net, the reference prompt $\mathbf{y}^r$ includes only the clothing-specific captions $\mathcal{K}^c$. In contrast, the main prompt $\mathbf{y}^m$ combines both the person-specific captions $\mathcal{K}^p$ and the clothing-specific captions $\mathcal{K}^c$. In practice, we use the following format as the main prompt: 
``a \textcolor{darkgreen}{\{body shape ($a_1^p$)\}} \textcolor{darkgreen}{\{gender ($a_2^p$)\}} wears \textcolor{lightblue}{\{cloth category ($a_1^c$)\}}, \textcolor{lightblue}{\{material ($a_2^c$)\}}, ..., with \textcolor{darkgreen}{\{hand pose ($a_n^p$)\}}.'', where green and blue color denote person and clothing attributes, respectively. Therefore, our approach can generate the main prompt for the resulting image, even when arbitrary clothing is provided.

Each prompt is processed through a text encoder before being input to its respective U-Net. Additional details on the exemplar dataset, task descriptions, and templates are provided in the supplementary material.

\subsection{Enhancing Adaptability via Mask Refinement}
\textbf{Random Dilation Mask Augmentation.}
We train a virtual try-on model via generative textual prompt that allows for text-based editing. However, we note the limitations in the commonly used masking approach, known as \textit{clothing-agnostic person representation}~\cite{choi2021viton}, frequently used in virtual try-on methods. While such a masking approach effectively preserves the original person's appearance, it also retains certain features from the original clothing such as length and fit.
This constrained mask region causes the reference clothing to fit too closely to the mask boundaries during training, making the new clothing mimic the shape of the original clothing and creating potential conflicts during manipulation.

To address these issues, we propose random dilation mask augmentation. As illustrated in Fig~\ref{fig:main}, we introduce a coarse mask $\mathbf{m}_c$ and a fine mask $\mathbf{m}_f$ to enable learning across a diverse range of masked images. We randomly dilate the fine mask, ensuring it does not extend beyond the boundaries of the coarse mask. The dilated mask $\mathbf{m}_d$ used for training is represented as follows:
\begin{equation}
    \mathbf{m}_d = (\mathbf{m}_f \oplus^n \mathbf{b}) \cap \mathbf{m}_c, 
\end{equation}
where $\oplus^n$ denotes $n$-iterated dilation with a structuring element $\mathbf{b}$~\cite{haralick1987image}, up to a sufficiently large but finite $n$.

\noindent\textbf{Prompt-aware Mask Generation.}
For inference, we introduce a novel coarse-to-fine generation to effectively preserve the original person’s appearance while allowing flexible text-based image manipulation. As shown in Fig.~\ref{fig:PMG}, we start with a coarse mask to create an initial approximation of the clothing region aligned with the text prompt. For efficiency, we apply early stopping in the denoising process, running it only from timestep $T$ to $\sigma T$, where $\sigma \in [0, 1)$. We then approximate $\hat{\mathbf{z}}_0$, decode it to $\hat{\mathbf{x}}_0$, and segment the region of interest using an off-the-shelf human parsing model. By taking the union of this mask with an existing clothing-agnostic person representation (fine mask), we acquire a refined mask, which subsequently serves as the inpainting mask for generating the final output. This approach balances precision and efficiency, improving the alignment with the text prompt.

\begin{figure*}[t!]
    \centering
    \includegraphics[width=1\linewidth]{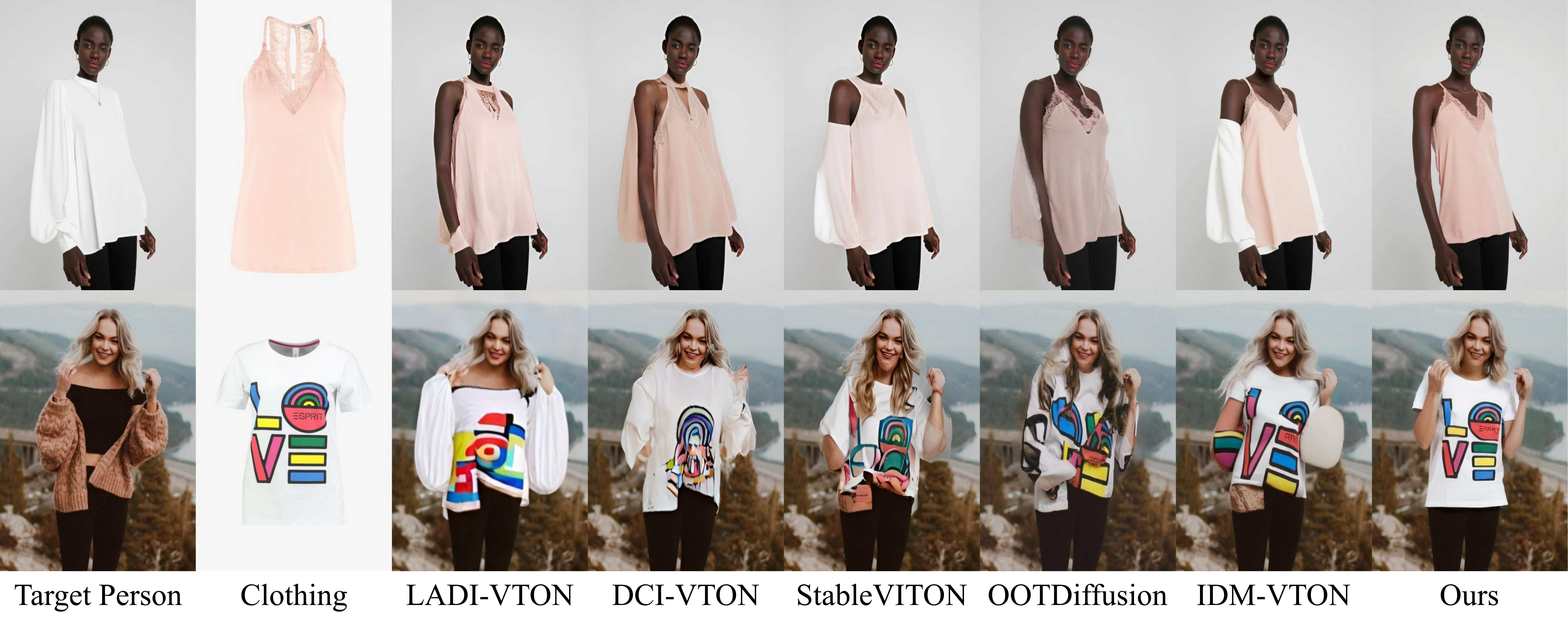}
    \caption{Qualitative comparison with baselines trained on VITON-HD dataset (first row: VITON-HD, second row: SHHQ-1.0).}
    \label{fig:vitonhd}
\end{figure*}

\begin{figure*}[t!]
    \centering
    \includegraphics[width=1\linewidth]{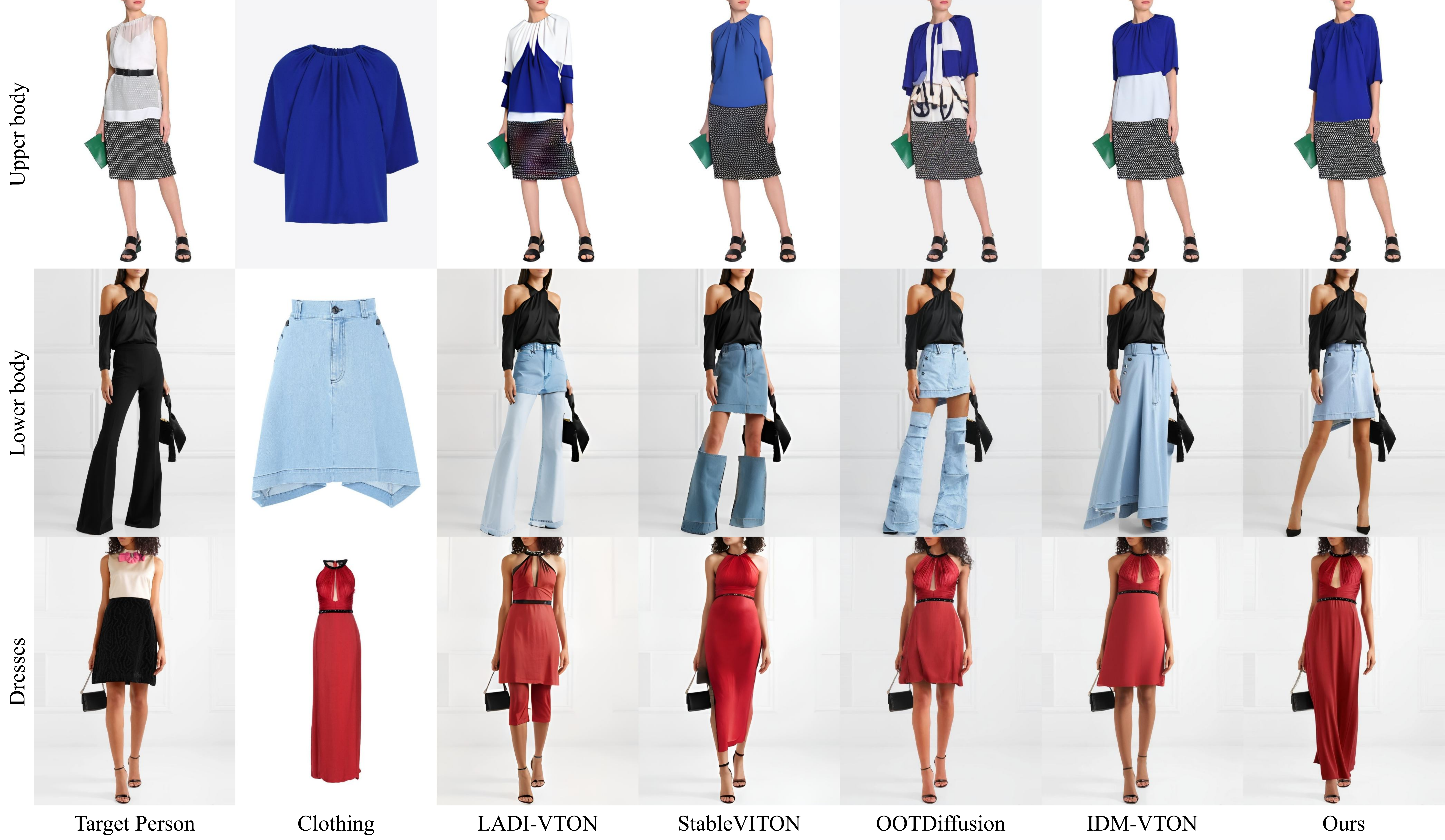}
    \caption{Qualitative comparison with baselines trained on DressCode dataset.}
    \label{fig:dresscode}
\end{figure*}

\section{Experiment}
We train~\sysname~separately on VITON-HD~\cite{choi2021viton} and DressCode~\cite{morelli2022dress} datasets, and SHHQ-1.0 is used to evaluate the generalizability of the model trained on VITON-HD~\cite{kim2024stableviton}. We provide details on the baseline and implementation in the supplementary material.

\subsection{Comparison with Baselines}\label{sec:quan}

\noindent\textbf{Qualitative Results.}
Fig.~\ref{fig:vitonhd} compares images generated by a model trained on VITON-HD, evaluating its performance on both the VITON-HD dataset (first row) and the SHHQ-1.0 dataset (second row).
The \textbf{baselines either closely follow the original clothing shape or lack detailed clothing features} (especially on SHHQ-1.0). In contrast, our model accurately captures the details and shape of the reference clothing while preserving the features of the original person image, such as background and pose. Additionally, we compared images of the upper body, lower body, and dresses on the DressCode~\cite{morelli2022dress} dataset in Fig.~\ref{fig:dresscode}. Across these images, \textbf{the baselines tend to follow to the original clothing shape.} Specifically in the lower body, baselines generate unnatural jeans by following the existing pants shape even when provided with a short skirt; for dresses, baselines also produce excessively short dresses.

\begin{table}[]
\centering
\resizebox{\linewidth}{!}{%
\begin{tabular}{lccccccc}
\toprule
Dataset & \multicolumn{4}{c}{VITON-HD} &  & \multicolumn{2}{c}{SHHQ-1.0} \\ \cline{2-5} \cline{7-8} 
Method & SSIM~$\uparrow$ & LPIPS~$\downarrow$ & FID~$\downarrow$ & KID~$\downarrow$ &  & FID~$\downarrow$ & KID~$\downarrow$ \\ 
\hline
LADI-VTON & 0.8630 & 0.1393 & 9.95 & 2.20 & & 26.23& 6.91 \\
DCI-VTON & 0.8712 & 0.1245 & 9.46 & 1.61 & & 26.39 & 7.37 \\
StableVITON & \underline{0.8757} & 0.1253 & 9.84 & 2.07 & & 23.97 & 7.30 \\
OOTDiffusion & 0.8424 & 0.1200 & 9.36 & \underline{1.04} & & 24.10 & \underline{6.20} \\
IDM-VTON & 0.8626 & \underline{0.1023} & 9.20 & 1.27 &  & 24.76 & 8.06 \\
IDM-VTON + our text & 0.8650 & 0.1025 & 9.15 & 1.26 &  & 23.78 & 7.46 \\
\hline
Ours\textsubscript{pose}  & \textbf{0.8778} & \textbf{0.0967} & \underline{9.07} & 1.16 &  & \underline{23.65} & 6.79 \\ 
Ours & 0.8686 & 0.1119 & \textbf{8.54} & \textbf{0.67} &  & \textbf{23.46} & \textbf{6.18} \\ 
\bottomrule
\end{tabular}%
}
\caption{Quantitative comparisons trained on VITON-HD. \textbf{Bold} and \underline{underline} denote the best and second best result, respectively.}
\label{Tab:main_vitonhd}
\end{table}

\begin{table*}[t!]
\centering
\resizebox{\linewidth}{!}{%
\begin{tabular}{llcccccccccccccc}
\toprule
Dataset &  & \multicolumn{4}{c}{Upper body} &  & \multicolumn{4}{c}{Lower body} &  & \multicolumn{4}{c}{Dresses} \\ \cline{3-6} \cline{8-11} \cline{13-16} 
Metric &  & SSIM~$\uparrow$ & LPIPS~$\downarrow$ & FID~$\downarrow$ & KID~$\downarrow$ &  & SSIM~$\uparrow$ & LPIPS~$\downarrow$ & FID~$\downarrow$ & KID~$\downarrow$ &  & SSIM~$\uparrow$ & LPIPS~$\downarrow$ & FID~$\downarrow$ & KID~$\downarrow$ \\ \hline
LADI-VTON &  & 0.9116 & 0.0995 & 15.06 & 3.78 &  & 0.8999 & 0.1072 & 14.78 & 2.97 &  & 0.8686 & 0.1348 & 14.70 & 3.35 \\
StableVITON &  & 0.9119 & 0.0925 & 14.13 & 3.34 &  & 0.8916 & 0.1104 & 14.91 & 3.10 &  & \underline{0.8745} & 0.1194 & 12.87 & 2.55 \\
OOTDiffusion &  & 0.9065 & 0.0755 & 15.04 & 2.96 &  & 0.8990 & 0.0741 & 14.05 & \underline{2.66} &  & 0.8554 & 0.1134 & 16.40 & 4.20 \\
IDM-VTON & \multicolumn{1}{c}{} & \underline{0.9267} & \textbf{0.0522} & \underline{11.91} & \underline{1.75} &  & \underline{0.9106} & \textbf{0.0608} & \underline{13.89} & 2.69 &  & \textbf{0.8787} & \textbf{0.0916} & \underline{11.36} & \underline{1.38} \\
Ours & \multicolumn{1}{c}{} & \textbf{0.9308} & \underline{0.0538} & \textbf{11.00} & \textbf{0.74} &  & \textbf{0.9165} & \underline{0.0642} & \textbf{12.55} & \textbf{1.46} &  & 0.8725 & \underline{0.0930} & \textbf{11.09} & \textbf{1.10} \\
\bottomrule
\end{tabular}%
}
\caption{Quantitative comparisons trained on DressCode dataset. \textbf{Bold} and \underline{underline} denote the best and the second best result, respectively.}
\label{Tab:main_dc}
\end{table*}

\begin{table}[t!]
\centering
\resizebox{0.7\columnwidth}{!}{%
\begin{tabular}{ccc}
\toprule
Methods & ``Untucked" & ``Tight fit" \\ \hline
Base Ratio & 44.64\% & 23.13\% \\
LADI-VTON & 50.78\% & 37.5\% \\
IDM-VTON & 46.31\% & 43.85\% \\
Ours\textsubscript{pose} & 62.84\% & 44.09\% \\
Ours & \textbf{89.42}\% & \textbf{66.98}\% \\
\bottomrule
\end{tabular}%
}
\caption{Evaluation on Text Alignment.}
\label{Tab:gpt_output}
\end{table}

\begin{table}[t!]
\centering
\resizebox{0.8\columnwidth}{!}{%
\begin{tabular}{lcccc} 
\toprule
Methods & SSIM$\uparrow$ & LPIPS$\downarrow$ & FID$\downarrow$ & KID$\downarrow$ \\
\hline
Ours w/ holi.  & 0.8574 & 0.1296 & 9.21 & 1.18 \\
Ours w/o PMG & 0.8411 & 0.1485 & 8.98 & 0.87 \\
Ours & \textbf{0.8686} & \textbf{0.1119} & \textbf{8.54} & \textbf{0.67}  \\ 
\bottomrule          
\end{tabular}%
}
\caption{Comparison of qualitative results with ablated methods.}
\label{Tab:abl}
\end{table}

\begin{figure}[t!]
    \centering
    \includegraphics[width=1\linewidth]{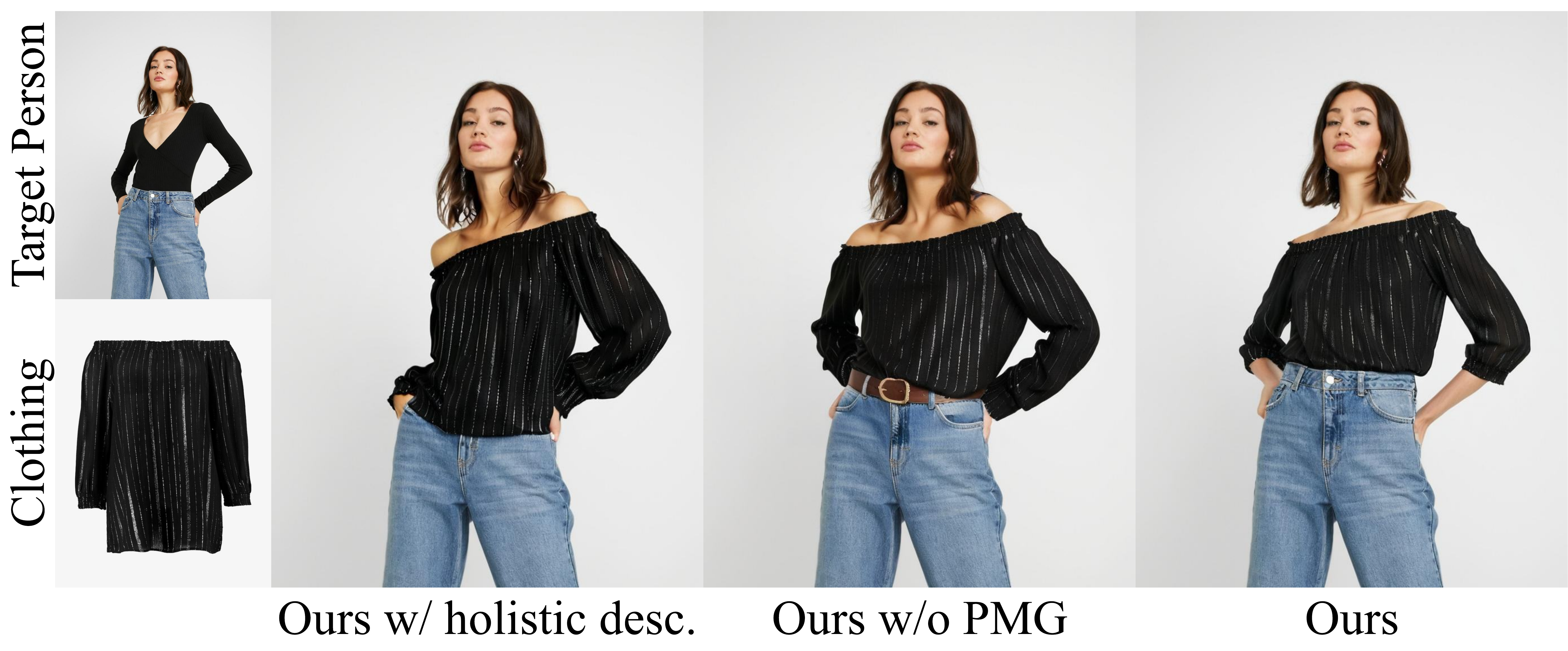}
    \caption{Visual comparisons for ablation studies.}
    \label{fig:abl}
\end{figure}

\noindent\textbf{Quantitative Results.}
For quantitative evaluation, we use four metrics: SSIM~\cite{detlefsen2022torchmetrics}, LPIPS~\cite{zhang2018unreasonable} (paired setting), and FID~\cite{heusel2017gans}, KID~\cite{binkowski2018demystifying} (unpaired setting). 
Tables~\ref{Tab:main_vitonhd} and~\ref{Tab:main_dc} show the comparison between~\sysname~and the baseline models on the VITON-HD and DressCode datasets, respectively. Our model consistently outperforms existing models in the unpaired setting (\textit{i.e.,} FID and KID). 
Notably, applying the generative textual prompt to IDM-VTON improved performance across all metrics except LPIPS, demonstrating the effectiveness of leveraging textual information for more realistic image generation.

While adding pose and hand descriptions slightly impacted performance in the paired setting due to minor pose misalignment, expanding the first convolution layer to 13 channels incorporating DensePose~\cite{kim2024stableviton} (Ours\textsubscript{pose}) shows the highest SSIM and LPIPS, as shown in Table~\ref{Tab:main_vitonhd}.
 However, performance dropped in the unpaired setting due to backbone modifications and pose-related errors, affecting generalization. Given minimal visual differences in qualitative and better unpaired results, we selected the model without spatial pose conditioning as the final choice.

\subsection{Further Analysis on~\sysname}
\noindent\textbf{Evaluation on Text Alignment.}
To validate our method's editing capability, we generated edited versions of 2,032 test images from VITON-HD by fixing a specific attribute (e.g., ``tucking style'') to a caption (\textit{e.g.}, ``untucked'') and then evaluated whether the captions generated by the LMM for these edited images matched the intended caption. 
We tested two settings: (i) setting the tucking style to ``untucked'' (\textit{i.e.}, top worn outside the pants) and (ii) ``tight fit''. 
The ``Base Ratio'' is the proportion of the 2,032 test images where the LMM identifies the caption. 
For example, if 1,016 images are identified as ``untucked,'' the base ratio for ``untucked'' is 50\%.
We compared our method with LADI-VTON, IDM-VTON, and Ours\textsubscript{pose}, all using text prompt.

Table~\ref{Tab:gpt_output} shows that our model achieved significantly higher accuracy for both attributes. Notably, the baseline models scored 50.78\% and 46.31\% for the ``untucked", performing similarly to the base ratio. This result underscores the limitations of conventional agnostic masks that restrict manipulation to pre-defined areas, especially for adjusting clothing length. Furthermore, for the ``tight fit" attribute, IDM-VTON and ours\textsubscript{pose} both achieved around 44\%, while our method reached 66.98\%, indicating that DensePose can hinder accurate text-based editing. 
These results show our approach outperforms conventional agnostic masks, enabling more precise manipulation from text prompts.

\noindent\textbf{Ablation Study.}
We investigate key design choices through ablation studies on the VITON-HD dataset, shown in Fig.\ref{fig:abl} and Table\ref{Tab:abl}. We compare: (i) using a holistic text description (\textit{i.e.,} a overall description of the image in detail) for both person and clothing images with the LMM, and (ii) excluding PMG during inference.

Fig.\ref{fig:abl} shows that a holistic description fails to capture complex poses accurately, and Table\ref{Tab:abl} reveals it results in the worst FID and KID scores. This indicates that LMM-based prompts without in-context learning miss fine details like poses.
Without PMG, the coarse masks often lead to inaccuracies (e.g., missing a belt, as shown in Fig.\ref{fig:abl}). Table\ref{Tab:abl} confirms that PMG significantly improves performance across all metrics.

\begin{figure}[t!]
    \centering
    \includegraphics[width=1\linewidth]{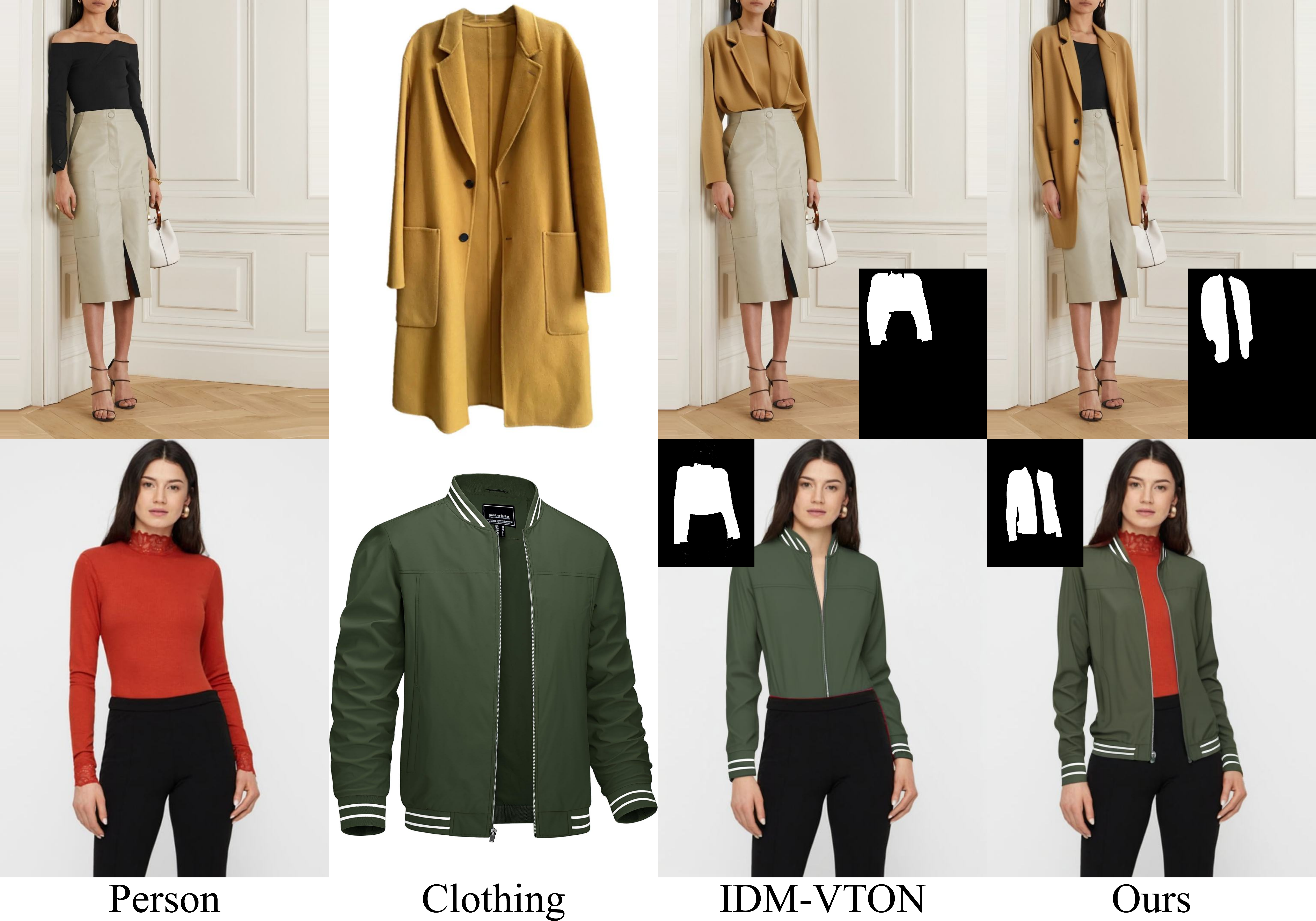}
    \vspace{-0.5cm}
    \caption{Extended Applications of~\sysname}
    \vspace{-0.5cm}
    \label{fig:effect_outer}
\end{figure}

\noindent\textbf{Extended Applications of~\sysname.}
In Figure~\ref{fig:effect_outer}, we further demonstrate our method’s effectiveness in layering outerwear. For instance, in the second row of of Fig.~\ref{fig:effect_outer}, both models were given a text: ``woman is wearing a outerwear with the zipper open.", and we use Grounded-SAM~\cite{ren2024grounded}.

IDM-VTON struggle to accurately interpret the text prompt and rely on constrained mask regions during training and inference, causing them to conform to the shape of the existing innerwear. In contrast, our approach, enhanced by rich text prompts and PMG, successfully preserves the red innerwear, highlighting PMG’s ability to achieve precise, text-driven clothing manipulation.

\begin{figure}[t!]
    \centering
    \includegraphics[width=1\linewidth]{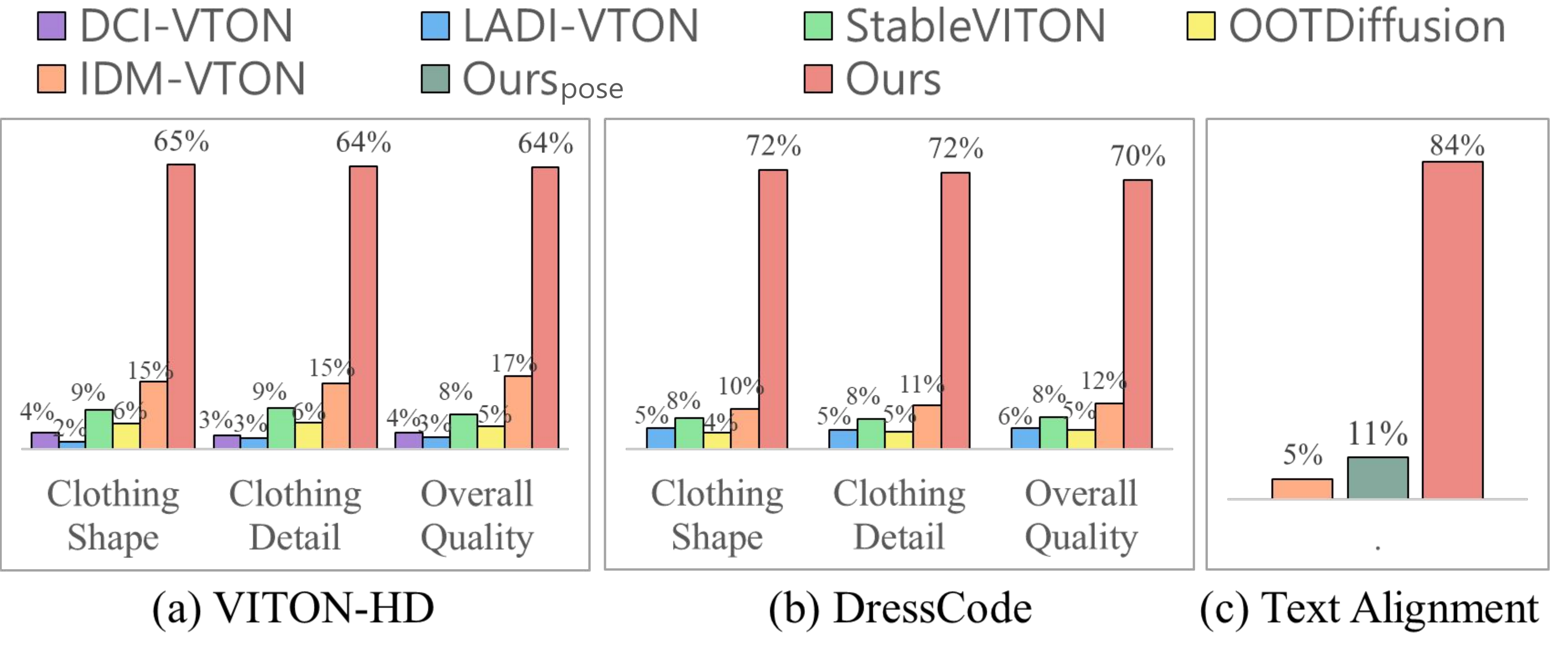}
    \caption{User Study Results. We requested users to identify the most appropriate samples for each criterion from (a) the VITON-HD, SHHQ, and (b) DressCode dataset. We performed (c) a qualitative assessment of text alignment using the VITON-HD.}
    \label{fig:user}
\end{figure}

\noindent\textbf{User Study}
We conducted a user study with 40 participants using the VITON-HD and DressCode datasets to evaluate our method and baselines. 
As shown in Fig.~\ref{fig:user} (a) and (b), we compared five models on VITON-HD and six models on DressCode, with participants rating results based on clothing shape, detail, and overall quality. 
Our method received significantly higher ratings across all three aspects.
For text editability, we compared our method with IDM-VTON (SDXL-based architecture) and Ours$_\text{pose}$ (with pose information). Participants were asked to select samples matching conditions like ``untucked", ``tight fit" and ``sleeve rolled up." 
As shown in Fig.~\ref{fig:user} (c), our model achieved 84\% preference, demonstrating superior text editability.

\noindent\textbf{Limitations and Future Work}
This paper demonstrated the utility of generative text prompts using LMMs with in-context learning.
However, due to SDXL’s 77-token limit, we manually constructed the in-context dataset.
Inspired by recent studies~\cite{li2024configure} on optimizing in-context sequences, exploring automated in-context data construction for virtual try-on is a promising direction. 
Stable Diffusion 3\cite{esser2024scaling}, with its T5 text encoder\cite{raffel2020exploring}, supports longer contexts, and leveraging well-configured in-context learning on such models could further enhance virtual try-on performance.

\section{Conclusion}
This paper introduces a novel virtual try-on model that leverages generative text prompt and advanced masking methods. Along with rich text descriptions from a large multimodal model, our approach not only improves performance but also enables versatile manipulation of virtual try-on results. Our dilated mask addresses the issue of generated clothing following too closely to the person’s existing clothing, and allows for a more natural overlay. We propose a prompt-aware mask generation technique, which enhances diversity while preserving the person’s original appearance. Our method achieves state-of-the-art results across multiple datasets, highlighting the effectiveness of generative textual prompt for virtual try-on.

\clearpage
\noindent\textbf{Acknowledgments}
This work was supported by Institute for Information \& communications Technology Planning \& Evaluation (IITP) grant funded by the Korea government (MSIT) (RS-2019-II190075, Artificial Intelligence Graduate School Program (KAIST)). This work was supported by the National Research Foundation of Korea (NRF) grant funded by the Korea government (MSIT) (No. RS-2025-00555621). This work was supported by Technological Innovation R\&D Program (RS-2024-00438252) funded by the Ministry of SMEs and Startups (MSS, Korea).

{
    \small
    \bibliographystyle{ieeenat_fullname}
    \bibliography{main}
}

\clearpage
\setcounter{page}{1}
\setcounter{section}{0}
\renewcommand{\thesection}{\Alph{section}}
\maketitlesupplementary

\section{Details of LMM-driven Virtual Try-on Captioning}
We provide detailed explanations of the exemplar datasets, task descriptions, and templates for the categories of upper body, lower body, and dresses in Fig.~\ref{fig:prompt_upper_body},~\ref{fig:prompt_lower_body}, and~\ref{fig:prompt_dresses}, respectively. 
We first gave the LMM model the instruction to identify and list detailed attributes of a given person image, including components, such as the facial expression, skin color, clothing logos. We then selected attributes related to the masked regions of the person image or associated with the style of wearing the clothing, such as pose, hair length, and tucking style. For clothing images, we excluded attributes describing fine details, such as logo shapes or patterns, but instead focused on high-level attributes, such as the clothing category or sleeve.

\section{Additional Details on User Study}
In our user study, we recruited 40 participants to evaluate the images generated by the baselines across 30 questions. For each question, participants selected the model that best addressed the specified criteria.

For questions 1-25, participants compared the images from multiple datasets. Questions 1–10 featured images from six models (\textit{i.e.}, DCI-VTON, LADI-VTON, StableVITON, OOTDiffusion, IDM-VTON, and Ours), based on the VITON-HD and SHHQ-1.0 datasets. Questions 11-25 include three categories of DressCode dataset: upper-body clothing (Questions 11-15), lower-body clothing (Questions 16-20), and dresses (Questions 21-25). For these questions, images from five models were compared, excluding DCI-VTON.

Participants answered the following three questions for each image set:
\begin{itemize}
    \item Clothing shape: Select the image that best reflects the length and shape of the given garment.
    \item Clothing details: Select the image that best reflects the text, texture, and pattern of the given garment.
    \item Overall quality: Select the image of the best overall quality.
\end{itemize}
For questions 26-30, participants evaluated images generated using the VITON-HD dataset and selected the one that best matched the style described as ``untucked, tight fit, and sleeve rolled up.”

\begin{table}[h!]
\centering
\resizebox{0.8\columnwidth}{!}{%
\begin{tabular}{lcccc} 
\toprule
Methods & SSIM$\uparrow$ & LPIPS$\downarrow$ & FID$\downarrow$ & KID$\downarrow$ \\
\hline
Ours w/ LLaVA  & 0.8663 & 0.1175 & 8.85 & 0.91 \\
Ours w/ GPT-4o & \textbf{0.8686} & \textbf{0.1119} & \textbf{8.54} & \textbf{0.67}  \\ 
\bottomrule
\end{tabular}%
}
\caption{Ablation Results on VITON-HD~\cite{choi2021viton} dataset.}
\label{Tab:llava}
\end{table}

\section{Experimental Details}
\noindent\textbf{Baselines.}  We compare our model to four diffusion-based models (LADI-VTON\cite{morelli2023ladi}, DCI-VTON~\cite{gou2023taming}, StableVITON~\cite{kim2024stableviton}, and IDM-VTON~\cite{choi2024improving}).
We use pre-trained weights if available; otherwise, we re-implement them using official code. LADI-VTON, DCI-VTON, and StableVITON, all based on Stable Diffusion 1.5, generate images at 512$\times$384 resolution. To ensure a fair comparison, we upscale the outputs to 2$\times$ using Real-ESRGAN~\cite{wang2021real}. 

\noindent\textbf{Implementation Details.} We utilize a frozen SDXL~\cite{podell2023sdxl} and SDXL inpainting model~\cite{diffusers2024stable} as the reference and main U-Net, respectively. 
During inference, we set the denoising step as 30 with $\sigma$ set to 0.5 for prompt-aware mask generation. To maintain overall pose consistency, we retain hand and foot details within the inpainting mask by Sapiens~\cite{khirodkar2025sapiens}. Additionally, we use GPT-4o~\cite{achiam2023gpt} to automatically generate high-quality captions for pre-defined attributes across all experimental datasets.

\section{Additional Experimental Results}
\noindent\textbf{Comparison to other LMMs.}
In this paper, we utilize GPT-4o to generate captions for all experimental datasets.
To investigate whether our model exhibits a high dependency on GPT-4o in test time, we evaluated it using text prompts generated by an open-source LMM called LLaVA~\cite{liu2024visual}. 
As shown in Table~\ref{Tab:llava}, prompts from LLaVA exhibit slightly degraded performances in the unpaired setting (\textit{i.e.,} FID and KID) but achieve comparable scores in a paired setting (\textit{e.g.,} SSIM and LPIPS), compared to GPT-4o. 
This demonstrates that the proposed textual prompt can effectively be generated by open-source LMMs such as LLaVA, other than GPT-4o.

\begin{table}[h!]
\centering
\resizebox{0.7\linewidth}{!}{%
\begin{tabular}{lcccc}
\toprule
$\sigma$ & SSIM~$\uparrow$ & LPIPS~$\downarrow$ & FID~$\downarrow$ & KID~$\downarrow$ \\ 
\hline
0.8 & 0.868 & 0.1122 & 8.60 & 0.68 \\
0.7 & 0.868 & 0.1119 & \textbf{8.53} & 0.69 \\
0.6 & 0.868 & 0.1120 & 8.55 & 0.71 \\
0.5 & \textbf{0.869} & 0.1119 & 8.54 & 0.67 \\
0.4 & \textbf{0.869} & \textbf{0.1118} & 8.54 & 0.69 \\
0.3 & \textbf{0.869} & \textbf{0.1118} & \textbf{8.53} & \textbf{0.65} \\
\bottomrule
\end{tabular}%
}
\caption{Ablation results for $\sigma$ values.}
\label{tab:supple_sigma}
\end{table}

\noindent\textbf{Ablation Study on $\sigma$.}
In this paper, we introduce a novel prompt-aware mask to preserve the original person's appearance and enable flexible text-based image manipulation. In generating the mask, we apply early stopping for computational efficiency and adjust the number of inference steps through a hyper-parameter $\sigma$. As the value of $\sigma$ increases, the generation time for the prompt-aware mask decreases. We set the number of denoising steps to 30 across all configurations. Table~\ref{tab:supple_sigma} shows the performance behavior based on different $\sigma$ values. The lowest $\sigma$ value (0.3) results in more accurate refined masks, achieving the best performance across all metrics. However, slightly reduced performance can be traded off for efficient inference times. In this paper, we adopt $\sigma=0.5$, which offers inference efficiency while maintaining FID and KID values comparable to those achieved with $\sigma=0.3$.

\begin{table}[h!]
\centering
\resizebox{\linewidth}{!}{%
\begin{tabular}{lccccc}
\toprule
Method & sec/image & SSIM~$\uparrow$ & LPIPS~$\downarrow$ & FID~$\downarrow$ & KID~$\downarrow$ \\ 
\hline
IDM-VTON (40step)        & 5.84 & 0.8613 & 0.1018 & 9.14 & 1.18 \\
Ours\textsubscript{pose} & 5.78 & \textbf{0.8778} & \textbf{0.0967} & 9.07 & 1.16 \\
Ours                     & 5.78 & 0.8686 & 0.1119 & \textbf{8.54} & \textbf{0.67} \\
\bottomrule
\end{tabular}%
}
\caption{Comparison of IDM-VTON at similar inference time.}
\label{tab:supple_inference_time}
\end{table}

\noindent\textbf{Comparison of Inference Time.}
PMG derives a refined mask from the coarse mask during the initial denoising steps. While this process may introduce additional computational overhead, Table~\ref{tab:supple_inference_time} shows that, when compared to IDM-VTON with slightly increased steps, our method demonstrates superior performance in the unpaired setting while maintaining comparable inference time.

\begin{table}[h!]
\centering
\resizebox{0.7\linewidth}{!}{%
\begin{tabular}{lcc} 
\toprule
& \textbf{GPT-Human} & \textbf{Human-Human}  \\ 
\hline
STS & 0.8622             & 0.8889                \\
\bottomrule
\end{tabular}
}
\caption{STS between GPT-Human and Human-Human pairs.}
\label{tab:supple_correlation}
\end{table}

\noindent\textbf{Bias in LMM Evaluation.}
Recent studies~\cite{zhang2023gpt,han2025video} validated LMMs for evaluation tasks such as image-to-text and multi-image-to-text alignment.
Similarly, we used an LMM to evaluate outfit labeling.
To verify the LMM-based evaluation, a user study (Fig. 8(c) in the main paper) assessed text-image alignment via human feedback, showing significant improvement over baselines.
To verify LMM reliability, four human annotators labeled 100 VITON-HD test images. 
Table~\ref{tab:supple_correlation} shows mean semantic textual similarity (STS) between GPT and human labels (GPT-Human), as well as between all human-labeled pairs (Human-Human). 
Comparable GPT-Human and Human-Human scores confirm GPT’s alignment with human judgment.

\begin{figure}[h!]
    \centering
    \includegraphics[width=0.9\linewidth]{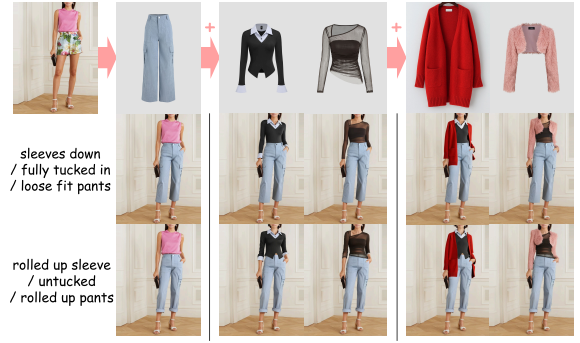}
    \caption{Multi-layer / transparent outfit generation images.}
    \label{fig:supple_multi_layer}
\end{figure}

\begin{table}[h!]
\centering
\resizebox{0.6\linewidth}{!}{%
\begin{tabular}{lcc} 
\toprule
& \textbf{SSIM}~$\downarrow$ & \textbf{LPIPS}~$\uparrow$  \\ 
\hline
\textbf{IDM-VTON} & 0.9401        & 0.0405          \\
\textbf{Ours}     & \textbf{0.8702}        & \textbf{0.1030}          \\
\bottomrule
\end{tabular}
}
\caption{Comparison of `tucked in' vs. `untucked'.}
\label{tab:supple_diversity}
\end{table}

\noindent\textbf{Additional Visual Results and Diversity.}
Fig.~\ref{fig:supple_multi_layer} demonstrates outfit changes and transparent cases generated under various textual conditions, combining complex scenarios such as tops, bottoms, and outerwear. Although trained solely on the DressCode dataset, our method effectively edits images with multi-layered clothing and diverse outfits. We plan to update the image combinations in future versions.
Table~\ref{tab:supple_diversity} compares SSIM and LPIPS between tucked and untucked generations. 
By leveraging rich text prompts and flexible mask, our method achieves higher editability than IDM-VTON.

\begin{figure}[h!]
    \centering
    \includegraphics[width=1.0\linewidth]{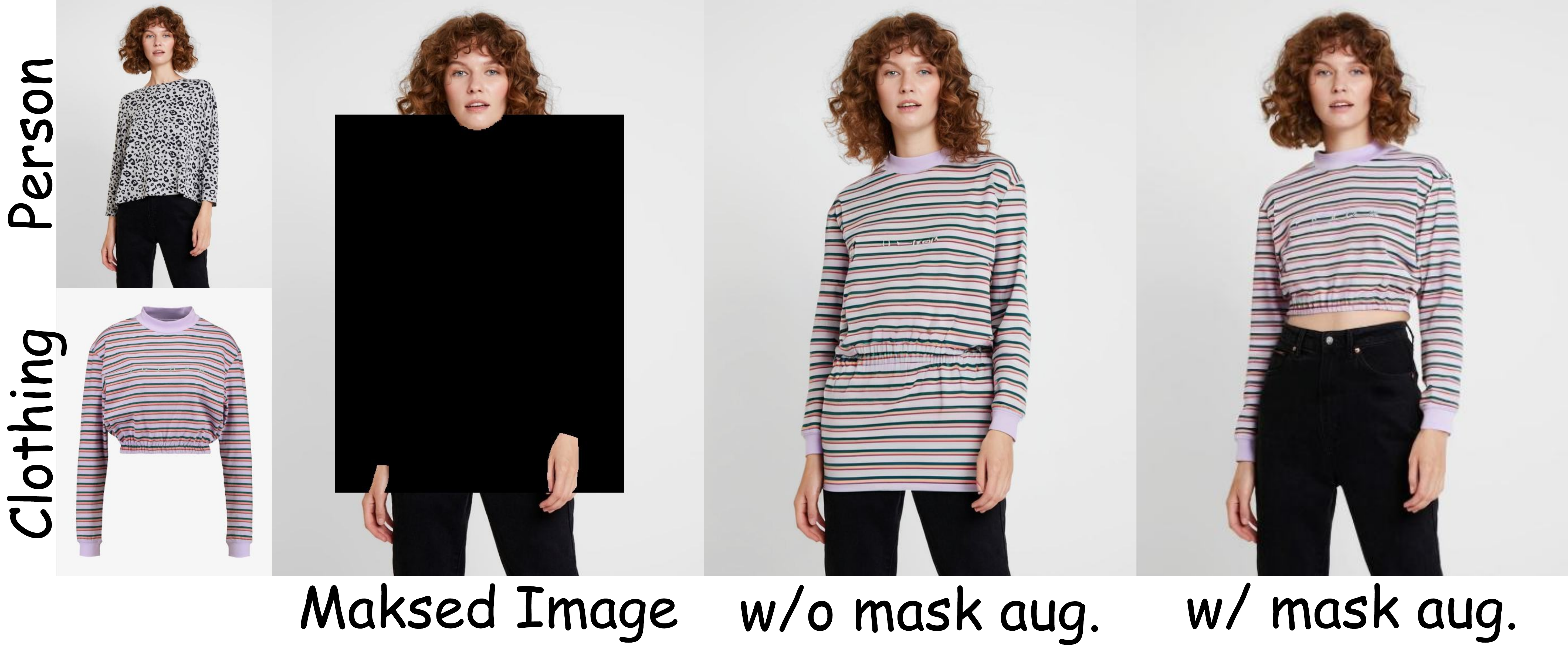}
    \caption{Generation results of VITON-HD.}
    \label{fig:supple_maskaug}
\end{figure}

\noindent\textbf{Additional Visual Results of Mask Augmentation. }
Fig.~\ref{fig:supple_maskaug} shows the effect of mask augmentation. As noted in the paper, training solely with fine masks causes the model to fit the clothing strictly within the mask region. Without augmentation, it fails to handle coarse masks (\textit{e.g.}, the large rectangular mask), generating overly long garments, as shown in Fig.~\ref{fig:supple_maskaug}. 

\noindent\textbf{Additional Qualitative Comparisons.}
We present additional qualitative results in Fig.~\ref{fig:supple_qual_vitonhd_shhq} and~\ref{fig:supple_qual_dc}. The first three rows in Fig.~\ref{fig:supple_qual_vitonhd_shhq} depict generated images on the VITON-HD~\cite{choi2021viton} dataset using a model trained on the same dataset, while the fourth and fifth rows show generated images on the SHHQ-1.0~\cite{fu2022styleganhuman} dataset. Our model consistently generates the most realistic images, even for complex poses (rows 1 and 2), and addresses the issue of following the shape of the original clothing (rows 3 and 4). Notably, in the third row, only our model accurately captures the shape of the given cropped top. Moreover, Fig.~\ref{fig:supple_qual_dc} shows additional results for the upper body, lower body, and dresses categories on the DressCode~\cite{morelli2022dress} dataset. Similar to the results on the VITON-HD dataset, our ~\sysname~accurately generate the length of the clothing and mitigate the constraint the model follows the original clothing's shape, highlighting the effectiveness of our rich text prompts and a novel mask refinement process.

\noindent\textbf{Additional text-based editing Results.}
Fig.~\ref{fig:supple_textedit_viton} and~\ref{fig:supple_textedit_dc} demonstrate the text-editing capability of our~\sysname~on VITON-HD and lower body category of the DressCode datasets, respectively. Fig.~\ref{fig:supple_textedit_viton} shows variations in tucking styles on the VITON-HD dataset, where the given clothing is generated based on the text prompts ``fully tucked in", ``untucked", and ``french tucked". Fig.~\ref{fig:supple_textedit_dc} presents variations on the DressCode dataset, including ``loose fit," ``tight fit," and ``pants rolled up." The generated results demonstrate accurate and text-based editing capability of~\sysname.

\begin{figure*}[t!]
    \centering
    \includegraphics[width=0.9\linewidth]{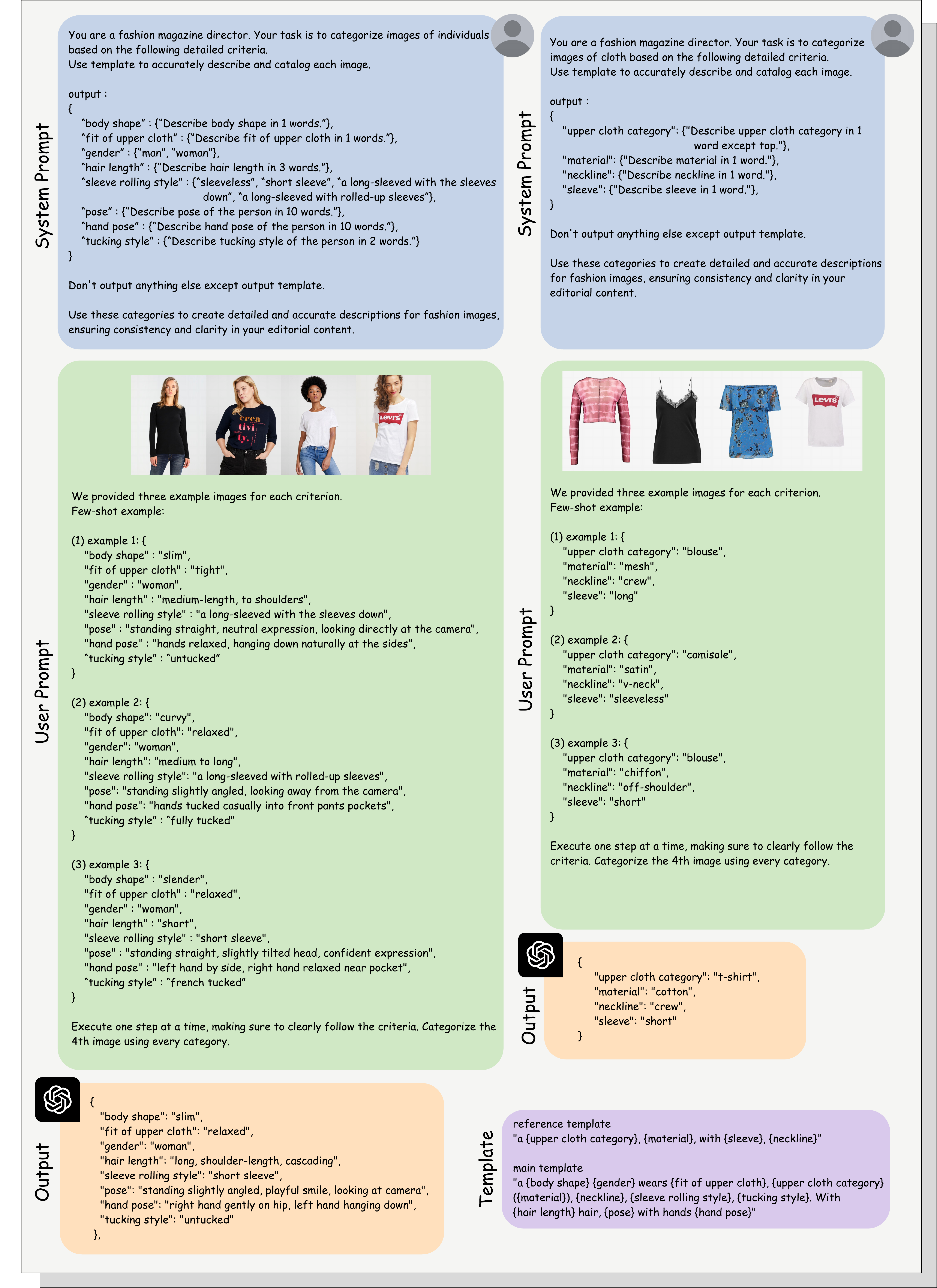}
    \caption{Detailed explanation of the exemplar dataset, task description, and templates for the upper body category.}
    \label{fig:prompt_upper_body}
\end{figure*}

\begin{figure*}[t!]
    \centering
    \includegraphics[width=0.9\linewidth]{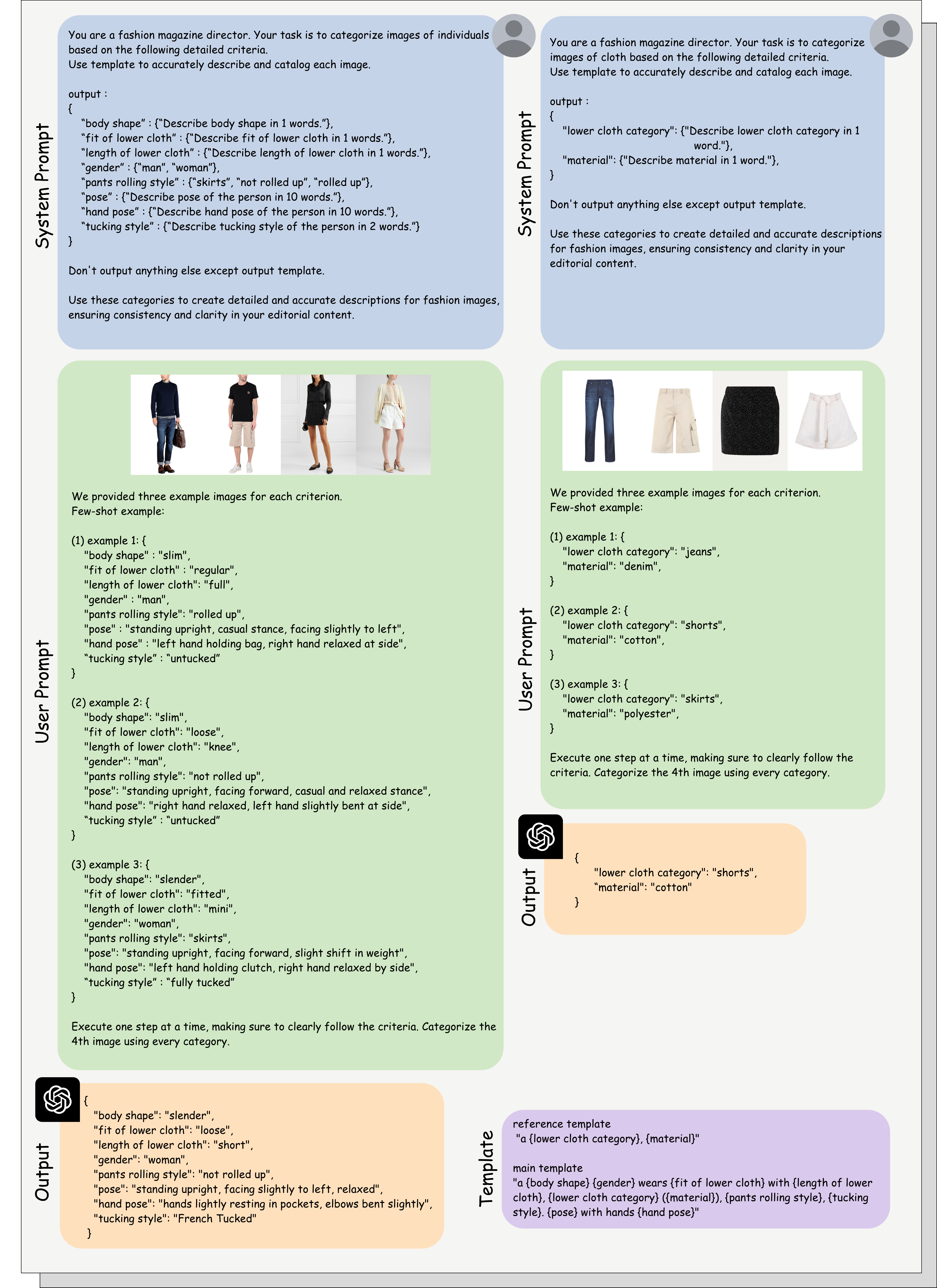}
    \caption{Detailed explanation of the exemplar dataset, task description, and templates for the lower body category.}
    \label{fig:prompt_lower_body}
\end{figure*}

\begin{figure*}[t!]
    \centering
    \includegraphics[width=0.9\linewidth]{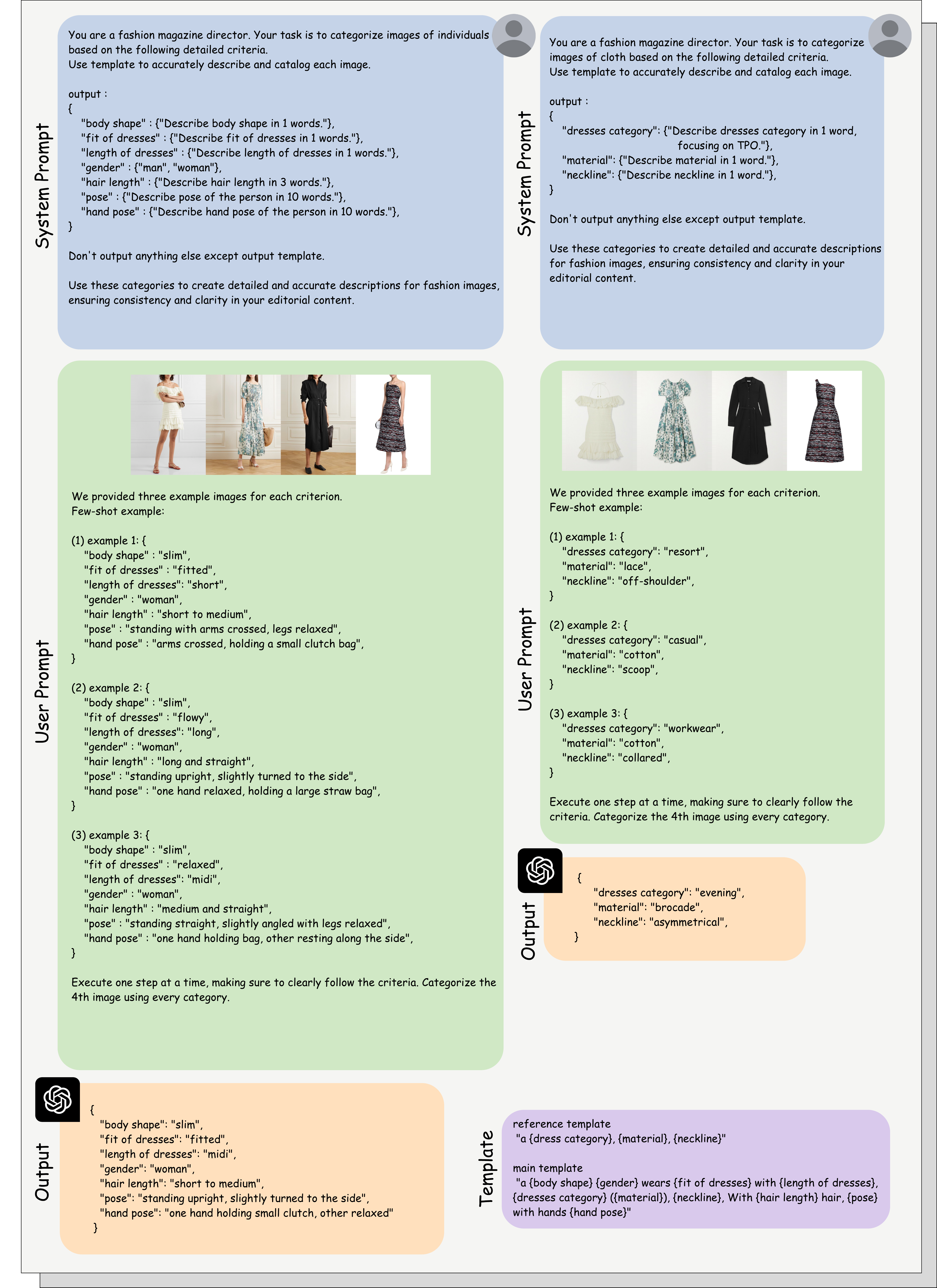}
    \caption{Detailed explanation of the exemplar dataset, task description, and templates for the dresses category.}
    \label{fig:prompt_dresses}
\end{figure*}

\begin{figure*}[t!]
    \centering
    \includegraphics[width=1\linewidth]{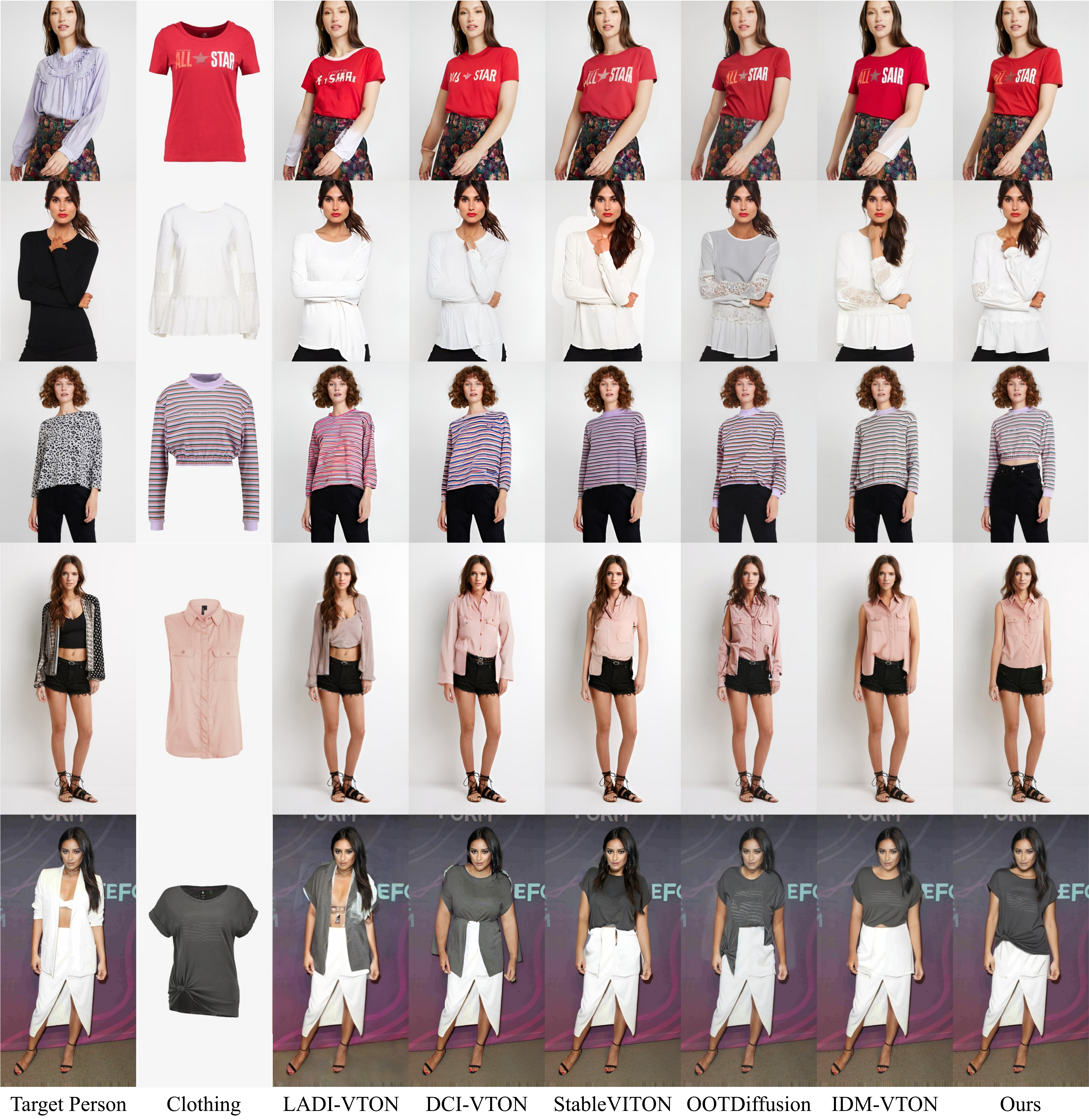}
    \caption{Qualitative comparison with baselines trained on VITON-HD dataset (first / second / third row: VITON-HD, fourth / fifth row: SHHQ-1.0)}
    \label{fig:supple_qual_vitonhd_shhq}
\end{figure*}

\begin{figure*}[t!]
    \centering
    \includegraphics[width=1\linewidth]{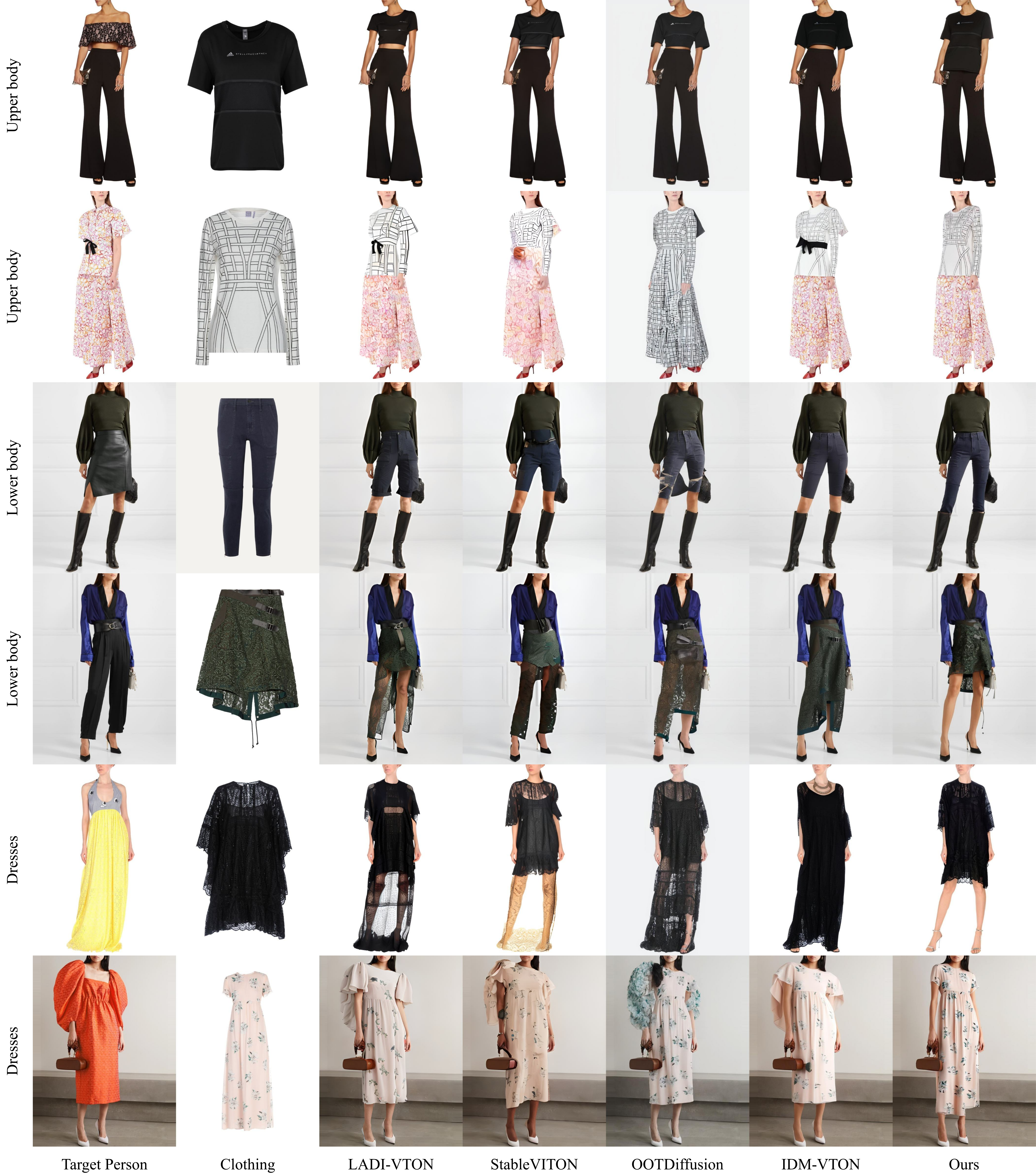}
    \caption{Qualitative comparison with baselines trained on DressCode dataset.}
    \label{fig:supple_qual_dc}
\end{figure*}

\begin{figure*}[t!]
    \centering
    \includegraphics[width=1\linewidth]{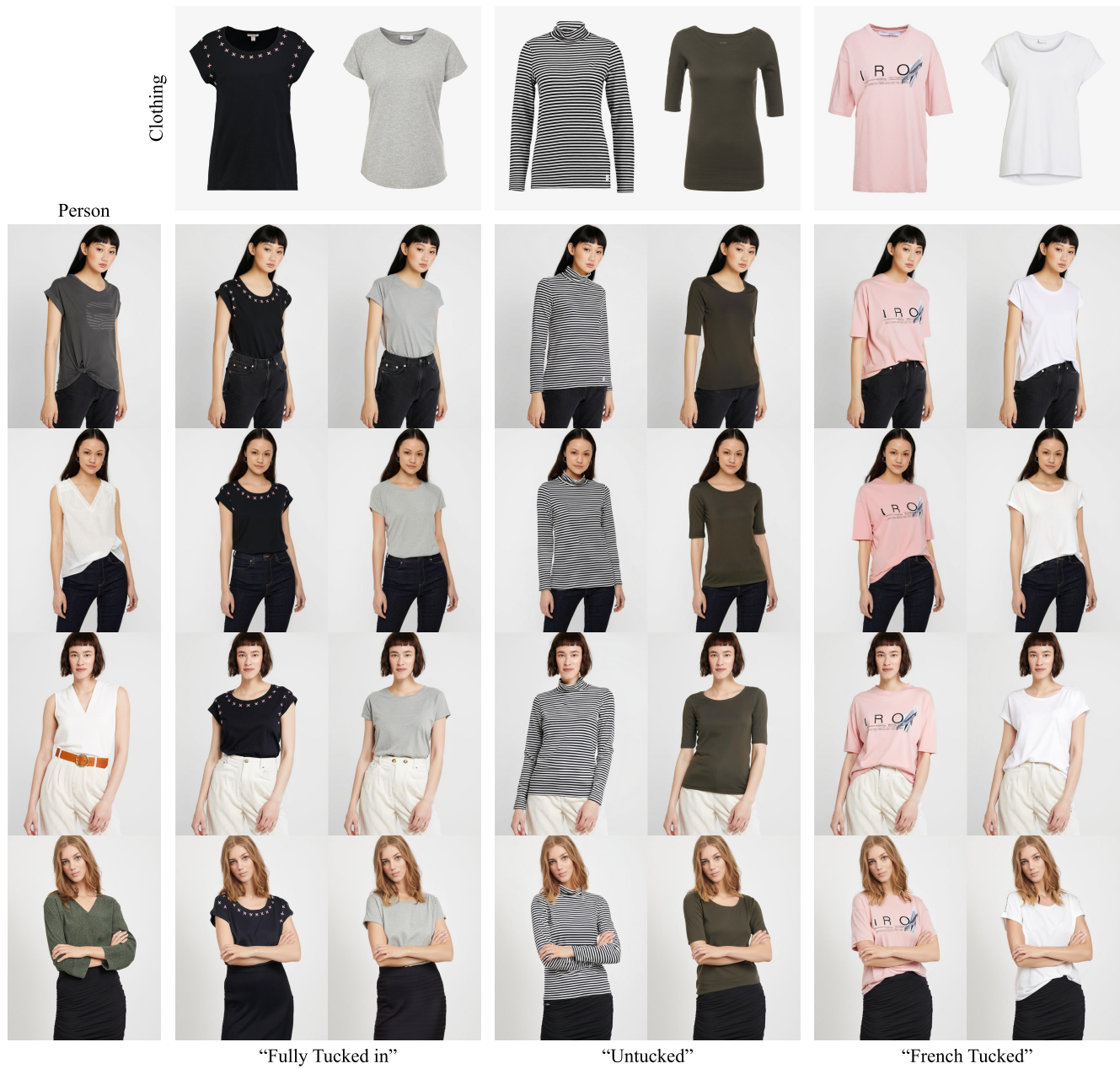}
    \caption{Additional text-based editing results for the upper body category of the VITON-HD dataset.}
    \label{fig:supple_textedit_viton}
\end{figure*}

\begin{figure*}[t!]
    \centering
    \includegraphics[width=1\linewidth]{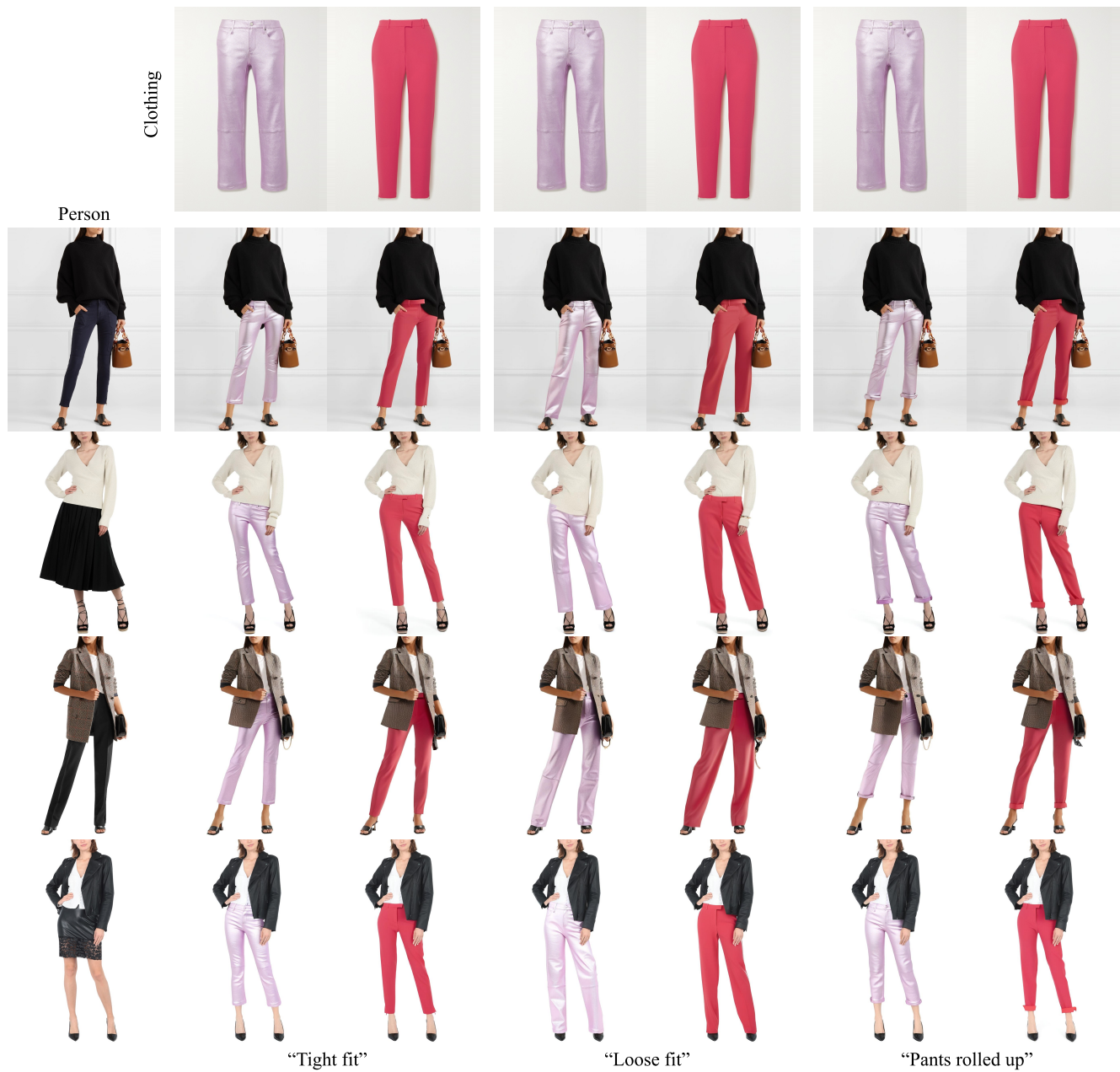}
    \caption{Additional text-based editing results for the lower body category of the DressCode dataset.}
    \label{fig:supple_textedit_dc}
\end{figure*}

\end{document}